\documentclass[letterpaper]{article} % DO NOT CHANGE THIS
\usepackage[preprint]{aaai2027}  % Public preprint mode
\usepackage[hyphens]{url}  % DO NOT CHANGE THIS
\usepackage{graphicx} % DO NOT CHANGE THIS
\usepackage{natbib}  % DO NOT CHANGE THIS AND DO NOT ADD ANY OPTIONS TO IT
\usepackage{caption} % DO NOT CHANGE THIS AND DO NOT ADD ANY OPTIONS TO IT
\usepackage{tabularx}
\usepackage{algorithm}
\usepackage{algorithmic}

\usepackage{newfloat}
\usepackage{listings}
\DeclareCaptionStyle{ruled}{labelfont=normalfont,labelsep=colon,strut=off} % DO NOT CHANGE THIS
\floatstyle{ruled}
\newfloat{listing}{tb}{lst}{}
\floatname{listing}{Listing}

\usepackage{booktabs}

\usepackage{amsmath}
\usepackage{amssymb}

\title{EEG-EditBench: Probing Visual Information in EEG--Image Retrieval Models with Controlled Image Edits}
\author{
    Kaifan Zhang\textsuperscript{\rm 1},
    Lihuo He\textsuperscript{\rm 1}\corresponding,
    Yuqi Ji\textsuperscript{\rm 1},
    Junjie Ke\textsuperscript{\rm 2},
    Lukun Wu\textsuperscript{\rm 1},
    Tianhao You\textsuperscript{\rm 3},
    Xinbo Gao\textsuperscript{\rm 1}
}
\affiliations{
    \textsuperscript{\rm 1}School of Electronic Engineering, Xidian University, Xi'an, China\\
    \textsuperscript{\rm 2}School of Software, Tsinghua University, Beijing, China\\
    \textsuperscript{\rm 3}Chongqing University of Posts and Telecommunications, Chongqing, China\\
    kaifanzhang@stu.xidian.edu.cn, lhhe@mail.xidian.edu.cn
}

\begin{document}

\maketitle

\begin{abstract}
Recent EEG-to-image retrieval models have achieved strong performance in identifying viewed images from semantically diverse candidates. Yet such success does not reveal what visual information supports the match. A model may readily identify a cheetah among tools, plants, and vehicles, but can it still distinguish the viewed cheetah from the same scene with the cheetah replaced by a dog? Motivated by this question, we introduce EEG-EditBench, a diagnostic benchmark that examines this question through controlled edits of object identity, attributes, background, and object presence. Built from the 200 THINGS-EEG2 test images, EEG-EditBench contains 2,137 quality-controlled edits and evaluates eight representative EEG visual decoding models. Our results show that strong standard retrieval does not consistently transfer to edit-based evaluation, with fine-grained attribute changes presenting the greatest challenge. EEG-EditBench reveals model behavior hidden by aggregate retrieval accuracy and provides a controlled basis for studying what visual information EEG--image models preserve. The code and complete dataset are available through the project repository.
\end{abstract}

\begin{links}
    \link{Code and Dataset}{https://github.com/XiaoZhangYES/EEG-EditBench}
\end{links}

\section{Introduction}

Decoding visual information from brain signals is a central problem in neuroscience and brain-computer interface research~\cite{kamitani2005decoding,kay2008identifying,naselaris2009bayesian,cichy2016comparison}. Large-scale natural-image datasets such as THINGS-EEG and THINGS-EEG2~\cite{grootswagers2022human,gifford2022large} have supported recent progress in EEG-to-image retrieval, where EEG responses and visual stimuli are mapped into a shared embedding space~\cite{song2024decoding,wei2024mb2c,li2024visual,wu2025bridging}. The standard evaluation is a 200-way zero-shot task that asks a model to retrieve the viewed image from candidates representing distinct object concepts. Recent methods have achieved strong performance under this protocol~\cite{jo2026hyfi,zheng2026hierarchical,liu2026blur}.

However, the standard 200-way protocol leaves an important question open. Consider an EEG response evoked by viewing an image of a cheetah. In the standard setting, the cheetah image is compared with candidates from different object concepts, such as a notebook, a drum, or an aircraft carrier. If the model retrieves the cheetah correctly, it has clearly learned a useful alignment between EEG and visual representations. Yet this success leaves open which visual distinctions the learned EEG--image alignment can support. The model may distinguish the target through relatively coarse cues such as ``animal,'' ``spotted animal,'' or ``animal in grassland,'' while finer distinctions in fur color, texture, background context, or nearby objects remain untested. Standard retrieval therefore shows that the model can find the viewed image among semantically diverse candidates, while leaving unclear which visual information supports the match.

\begin{figure*}[t]
\centering
\includegraphics[width=\textwidth]{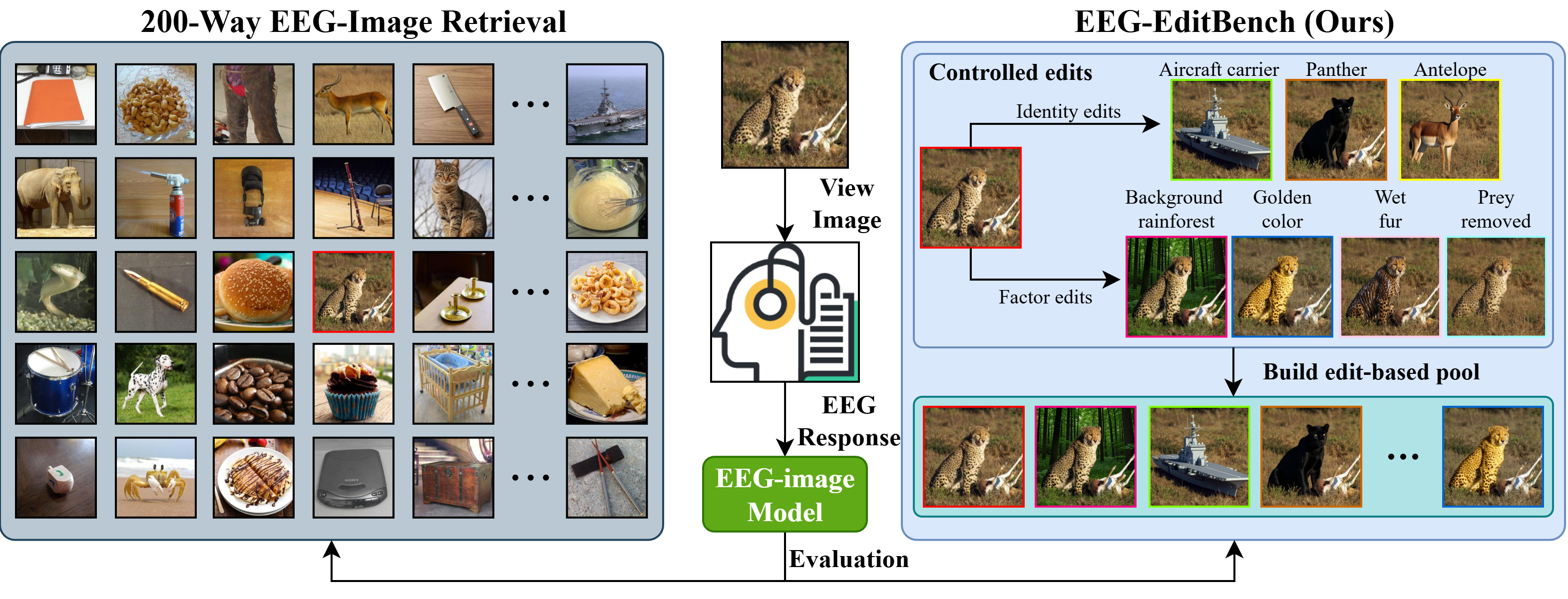}
\caption{From standard 200-way retrieval to edit-based diagnosis. Standard EEG-image retrieval asks whether a model can identify the viewed image from semantically diverse candidates. EEG-EditBench instead compares each viewed image with controlled variants, enabling factor-level analysis of EEG--image matching.}
\label{fig:teaser}
\end{figure*}

A single aggregate accuracy cannot resolve this ambiguity. Two models may achieve similar retrieval accuracy while relying on different visual evidence, and a model that performs well on cross-concept candidates may still fail when the alternatives differ only in a specific attribute or contextual detail. \textbf{Diagnostic evaluation} addresses this limitation through controlled examples that isolate particular model capabilities~\cite{ribeiro2020beyond,zhao2022vlchecklist,thrush2022winoground,hsieh2023sugarcrepe}. Image editing provides a practical way to construct such examples by changing a specific visual factor while preserving the remaining image content as much as possible~\cite{zhang2023magicbrush,ma2024i2ebench}. This perspective naturally extends to EEG--image retrieval, where the EEG query can remain fixed while the candidate image is varied in a controlled manner.

Building on this idea, we introduce \textbf{EEG-EditBench}, a controlled edit-based benchmark for probing visual information in EEG--image retrieval models. As illustrated in Figure~\ref{fig:teaser}, EEG-EditBench shifts evaluation from cross-concept retrieval to controlled comparisons between each source image and variants that alter object identity, object attributes, background context, or object presence. Starting from the 200 THINGS-EEG2 test images, we construct quality-controlled edited variants that retain substantial source-image content while differing in a specific visual factor.

EEG-EditBench evaluates whether a model can distinguish each viewed image from its controlled variants, both individually and when multiple edited variants compete within the same candidate pool. These complementary settings support factor-level analysis as well as a more challenging joint comparison.

We evaluate eight representative EEG visual decoding models on EEG-EditBench~\cite{song2024decoding,li2024visual,wei2024mb2c,wu2025bridging,wu2026shrinking,jo2026hyfi,zhang2026neurobridge,zheng2026hierarchical} and obtain three main findings:

\begin{itemize}
\item \textbf{High accuracy in standard EEG-to-image retrieval does not guarantee that a model can distinguish the viewed image from its controlled edits.}
\item \textbf{Across the evaluated models, object identity and removal edits are distinguished more accurately than edits to color, material, texture, shape, and state.}
\item \textbf{The same type of edit can differ in difficulty across image categories.}
\end{itemize}

\begin{figure*}[t]
    \centering
    \includegraphics[width=\linewidth]{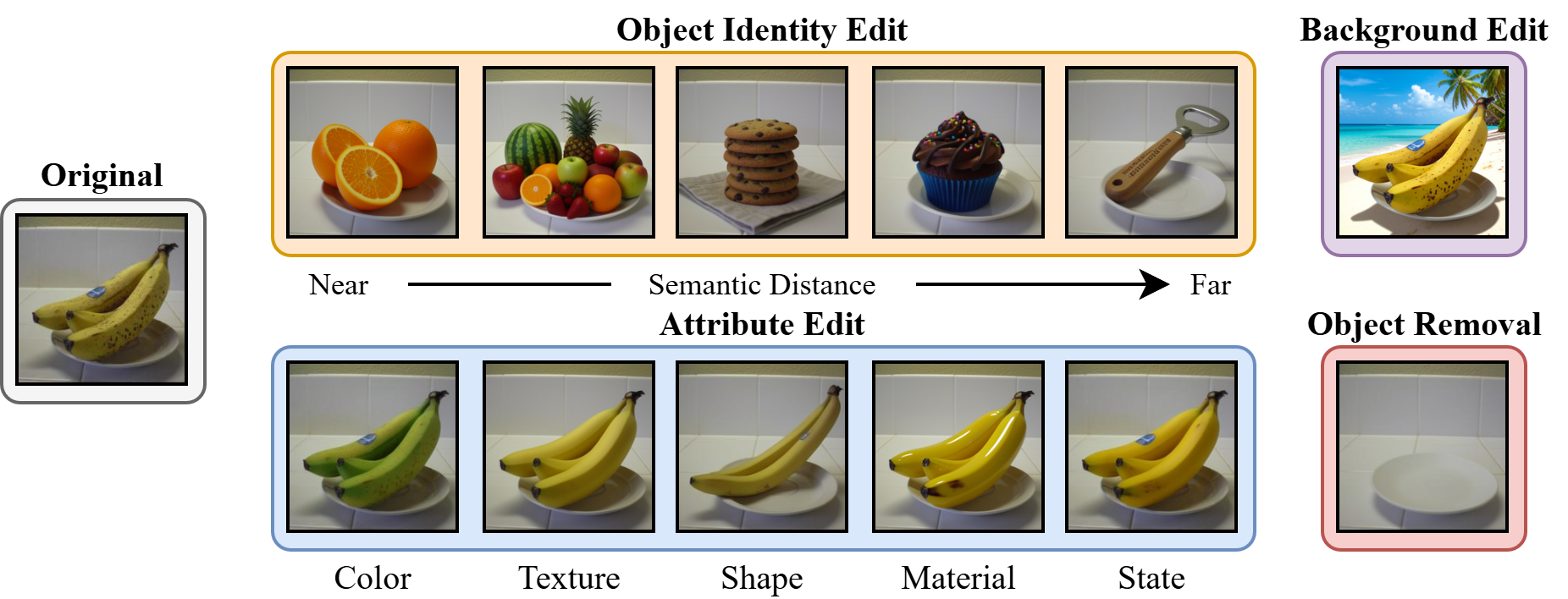}
    \caption{Edit taxonomy of EEG-EditBench. Identity edits are organized by semantic proximity and attribute edits by the changed visual property; background and removal edits target scene context and object presence, respectively.}
    \label{fig:edit_taxonomy}
\end{figure*}

\begin{figure*}[t]
    \centering
    \includegraphics[width=\linewidth]{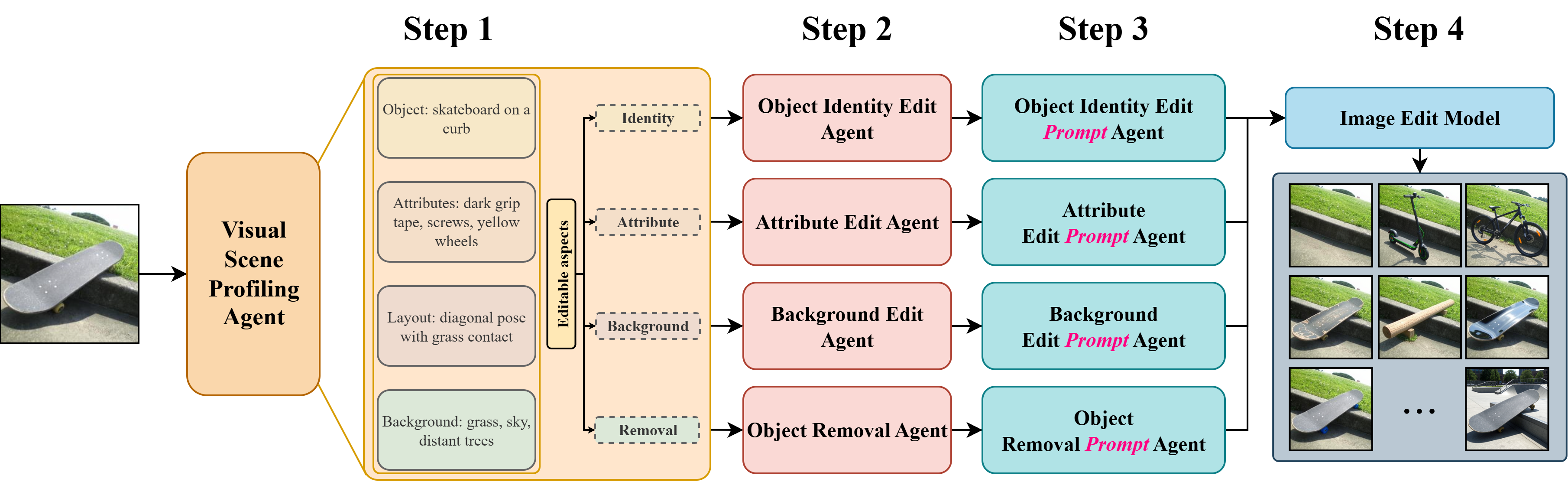}
    \caption{Dataset construction pipeline of EEG-EditBench. A structured scene profile guides edit-target and prompt generation through family-specific branches, followed by image editing.}
    \label{fig:dataset_pipeline}
\end{figure*}

\begin{figure}[t]
    \centering
    \includegraphics[width=\linewidth]{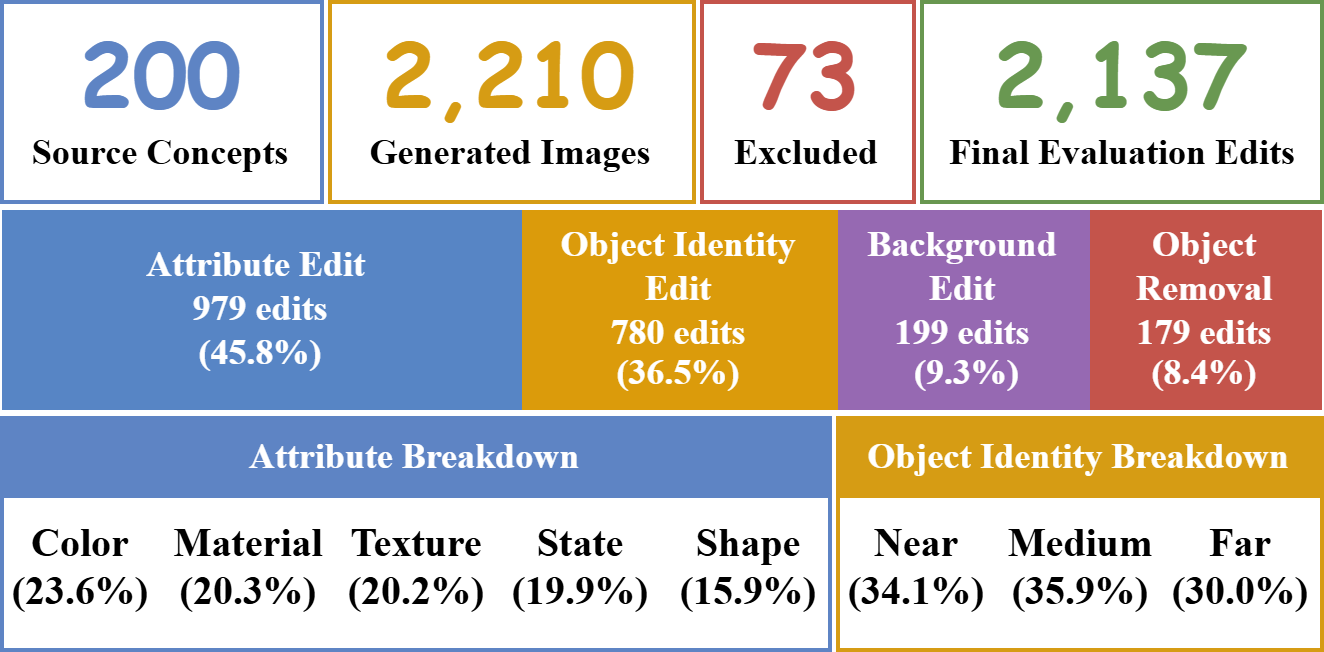}
    \caption{Composition of EEG-EditBench after quality control. The figure summarizes the dataset scale and the distribution of edit families and subtypes. Subtype percentages are computed within each edit family.}
    \label{fig:dataset_composition}
\end{figure}

\section{Related Work}

\subsection{EEG-Based Visual Decoding}

Earlier studies used single-trial EEG decoding and representational analysis to characterize the temporal dynamics of object processing, and explored multimodal and deep visual representation learning~\cite{kaneshiro2015representational,du2023decoding,singh2024learning}. Large-scale natural-image datasets such as THINGS-EEG and THINGS-EEG2 have since supported progress in visual decoding from EEG~\cite{grootswagers2022human,gifford2022large}. Existing methods either align EEG and visual representations for image retrieval~\cite{song2024decoding,wei2024mb2c} or combine EEG-derived representations with generative models for image reconstruction~\cite{bai2024dreamdiffusion,li2024visual, zhang2025cognitioncapturer, zhang2026cognitioncapturerpro}. More recent work improves EEG--image alignment through visual priors, hierarchical representations, hyperbolic modeling, adaptive teaching, and self-supervised objectives~\cite{wu2025bridging,zheng2026hierarchical,jo2026hyfi,wu2026shrinking,zhang2026neurobridge}.

These advances are commonly evaluated through aggregate retrieval or reconstruction metrics, which measure whether a model recovers the target image but reveal less about the visual information it preserves. Recent work has broadened brain-decoding evaluation toward fine-grained semantic content, multigranular image comparison, and mental imagery~\cite{xia2025exploring,xia2026multigranular,kneeland2025nsdimagery}.

\subsection{Diagnostic Evaluation with Controlled Image Edits}

Diagnostic benchmarks use controlled examples or hard negatives to expose model capabilities hidden by aggregate performance. This approach has been applied to compositional reasoning, language understanding, and vision-language grounding~\cite{johnson2017clevr,ribeiro2020beyond,shekhar2017foil,parcalabescu2022valse,zhao2022vlchecklist,thrush2022winoground,yuksekgonul2023when,ma2023crepe,hsieh2023sugarcrepe}. EEG-based retrieval differs because the query is a neural response evoked by viewing an image; factor-level diagnosis therefore varies the image candidates while keeping the EEG query fixed.

Image editing benchmarks provide a practical basis for this construction. Existing benchmarks evaluate instruction following, visual realism, and preservation of content unrelated to the requested change~\cite{wang2023imagen,zhang2023magicbrush,ma2024i2ebench}. Recent work expands edit diversity and preservation-oriented evaluation through inversion-based benchmarks, curated high-quality instruction--edit pairs, and unified multi-type editing datasets~\cite{ju2024pnp,hui2025hqedit,yu2025anyedit}. Counterfactual image generation similarly emphasizes targeted interventions with minimal unintended changes~\cite{melistas2024benchmarking}. EEG-EditBench follows these principles and uses quality-controlled edits as controlled candidates for measuring changes in EEG--image matching.

\section{Method}

\subsection{Problem Formulation}

For each concept \(c\), let \(e_c\) denote the EEG response evoked by the original source image \(I_c^{\mathrm{orig}}\). EEG-EditBench evaluates the same EEG query against a set of valid edited images \(E_c\), where \(E_c^{(f)} \subseteq E_c\) contains edits from edit family \(f \in \{\mathrm{id}, \mathrm{attr}, \mathrm{bg}, \mathrm{rm}\}\), corresponding to object identity, object attribute, background, and object removal.

For a given retrieval model, the EEG encoder \(\phi(\cdot)\) and image encoder \(\psi(\cdot)\) map EEG responses and images into a shared embedding space. Their similarity is written as
\[
s(e,I)=\langle \phi(e),\psi(I)\rangle,
\]
where the inner product denotes the model-specific score used for ranking. EEG-EditBench leaves this scoring function unchanged and modifies only the candidate images used for evaluation. It therefore characterizes how the complete EEG--image retrieval system behaves under controlled image-side changes, reflecting the combined effects of the EEG encoder, image encoder, and alignment objective. 

\subsection{Edit Taxonomy}

As illustrated in Figure~\ref{fig:edit_taxonomy}, EEG-EditBench defines four edit families that target complementary visual factors.

\paragraph{Object Identity Edit.}
This edit replaces the main object while preserving the composition, viewpoint, background, and visual style. Replacement targets are selected from the remaining THINGS-EEG2 test concepts and grouped into three scene-conditioned subtypes based on semantic distance: near targets are semantically related to the source or share a similar function or affordance; medium targets are less closely related but can still play a similar role in the scene; and far targets are semantically distinct yet remain plausible replacements.

\paragraph{Attribute Edit.}
This edit preserves the object identity and background while changing one visible property of the main object. We consider color, texture, shape, material, and state.

\paragraph{Background Edit.}
This edit replaces the surrounding scene while preserving the foreground object, its appearance, and its spatial arrangement.

\paragraph{Object Removal.}
This edit removes the main object and completes the exposed region with plausible background content, including associated shadows, reflections, or contact boundaries when present.

Together, the four edit families evaluate model behavior under changes in object identity, appearance, scene context, and object presence.

\subsection{Dataset Construction Pipeline}

EEG-EditBench is constructed from the 200 THINGS-EEG2 test images, whose object concepts originate from the THINGS database~\cite{hebart2019things,gifford2022large}, using the four-stage pipeline shown in Figure~\ref{fig:dataset_pipeline}. A vision-language model first produces a structured scene profile describing the main object, visible attributes, spatial layout, and background context, and determines the applicable edit families. It then generates edit targets and self-contained prompts that specify the intended change and the content to preserve. An image editing model uses each source image and prompt to generate the edited variant.

All generated images undergo full human review by two independent reviewers. Each image is assessed for whether the requested edit is correctly realized, whether content unrelated to the edit is sufficiently preserved, and whether the result remains visually coherent without salient editing artifacts. Images that fail any of these criteria are excluded, and disagreements between the two reviewers are resolved by a third reviewer. This process yields 2,137 retained edits. The review criteria and adjudication procedure are provided in the supplementary material.

The final benchmark contains 979 Attribute Edits, 780 Object Identity Edits, 199 Background Edits, and 179 Object Removals. Figure~\ref{fig:dataset_composition} shows the distribution across edit families and subtypes.

\begin{figure}[t]
    \centering
    \includegraphics[width=\linewidth]{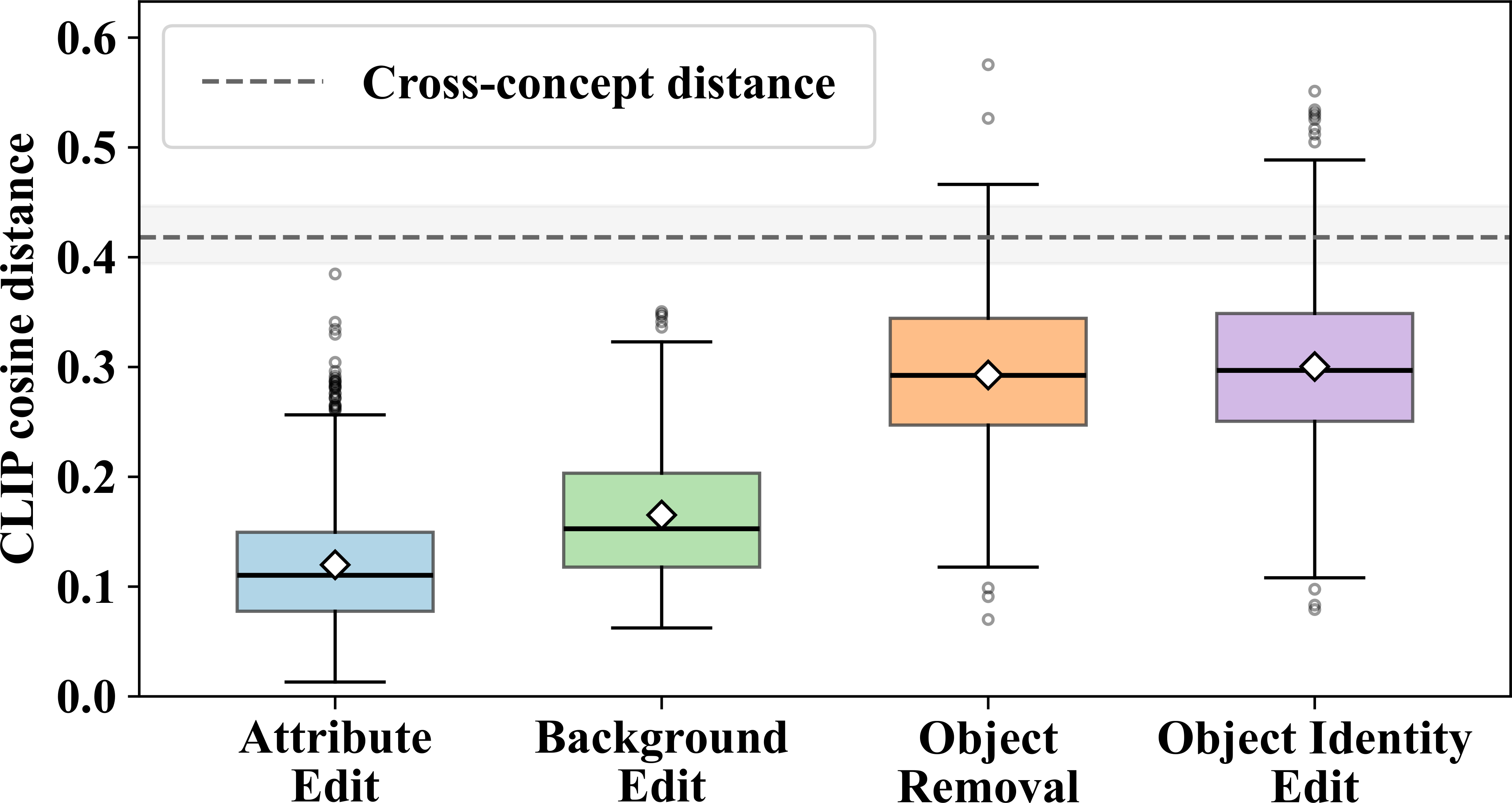}
    \caption{Cosine distances between edited images and their corresponding source images. The dashed line and shaded region show the median and interquartile range of the per-image median distances to source images from the other test concepts. Diamonds indicate means.}
    \label{fig:semantic_proximity}
\end{figure}

\begin{table*}[t]
\centering
\small
\setlength{\tabcolsep}{2.2pt}
\begin{tabular}{lcccccccccc}
\toprule
&
&
& \multicolumn{3}{c}{Retrieval}
& \multicolumn{5}{c}{2AFC Accuracy} \\
\cmidrule(lr){4-6} \cmidrule(lr){7-11}
Model
& Venue
& \shortstack{Visual\\Degradation}
& \shortstack{200-way\\Original}
& \shortstack{200-way\\Null}
& EP-Top1
& Identity
& Attribute
& Background
& Removal
& Overall \\
\midrule
NICE
& ICLR'24
& --
& 19.5 $\pm$ 3.8
& 19.8 $\pm$ 4.2
& 14.0 $\pm$ 2.6
& 79.1 $\pm$ 1.8
& 55.8 $\pm$ 3.5
& 66.6 $\pm$ 5.4
& 84.7 $\pm$ 3.7
& 67.7 $\pm$ 2.3 \\

ATM
& NeurIPS'24
& --
& 29.8 $\pm$ 5.5
& 22.7 $\pm$ 4.4
& 38.1 $\pm$ 5.9
& 86.7 $\pm$ 1.8
& 77.7 $\pm$ 4.1
& 77.8 $\pm$ 3.6
& 86.0 $\pm$ 3.7
& 81.7 $\pm$ 2.7 \\

MB2C
& ACM MM'24
& --
& 27.5 $\pm$ 6.3
& 25.7 $\pm$ 5.3
& 26.7 $\pm$ 3.2
& 83.5 $\pm$ 2.5
& 66.8 $\pm$ 3.1
& 77.1 $\pm$ 3.7
& 83.6 $\pm$ 3.0
& 75.3 $\pm$ 2.5 \\

UBP
& CVPR'25
& \checkmark
& 52.8 $\pm$ 5.3
& 46.7 $\pm$ 5.6
& 18.2 $\pm$ 2.2
& 88.8 $\pm$ 1.8
& 65.6 $\pm$ 2.1
& 80.4 $\pm$ 2.7
& 87.9 $\pm$ 3.9
& 77.3 $\pm$ 1.8 \\

ATS
& AAAI'26
& \checkmark
& 56.1 $\pm$ 7.3
& 49.5 $\pm$ 6.4
& 26.7 $\pm$ 3.2
& 89.7 $\pm$ 1.7
& 71.2 $\pm$ 2.1
& 84.8 $\pm$ 2.7
& 94.2 $\pm$ 2.3
& 81.1 $\pm$ 1.6 \\

NeuroBridge
& AAAI'26
& \checkmark
& 62.3 $\pm$ 6.5
& 59.8 $\pm$ 7.6
& 27.6 $\pm$ 2.5
& 91.0 $\pm$ 1.1
& 72.4 $\pm$ 1.9
& 87.2 $\pm$ 1.9
& 96.2 $\pm$ 1.1
& 82.6 $\pm$ 1.1 \\

HyFI
& AAAI'26
& \checkmark
& 68.2 $\pm$ 6.5
& 64.2 $\pm$ 7.0
& 33.4 $\pm$ 2.6
& 92.8 $\pm$ 1.1
& 75.7 $\pm$ 1.4
& 88.3 $\pm$ 2.3
& \textbf{97.7 $\boldsymbol{\pm}$ 1.5}
& 84.9 $\pm$ 1.1 \\

Brain-HIVE
& ICLR'26
& --
& \textbf{75.2 $\boldsymbol{\pm}$ 6.5}
& \textbf{65.5 $\boldsymbol{\pm}$ 5.7}
& \textbf{46.3 $\boldsymbol{\pm}$ 5.7}
& \textbf{95.6 $\boldsymbol{\pm}$ 1.3}
& \textbf{80.5 $\boldsymbol{\pm}$ 3.6}
& \textbf{92.4 $\boldsymbol{\pm}$ 2.7}
& 97.5 $\pm$ 1.2
& \textbf{88.6 $\boldsymbol{\pm}$ 2.2} \\
\bottomrule
\end{tabular}
\caption{Main results on standard retrieval and EEG-EditBench. Original and Null denote 200-way Top-1 accuracy using the original and null-edited candidate sets, respectively; EP-Top1 and 2AFC report edit-based performance. A checkmark indicates the use of degraded visual representations. Best results are shown in bold.}
\label{tab:main_results}
\end{table*}

\begin{table*}[t]
\centering
\small
\setlength{\tabcolsep}{2.0pt}
\begin{tabular}{lcccccccccc}
\toprule
& \multicolumn{4}{c}{Object Identity}
& \multicolumn{6}{c}{Attribute} \\
\cmidrule(lr){2-5} \cmidrule(lr){6-11}
Model
& Near
& Medium
& Far
& Overall
& Color
& Material
& Texture
& Shape
& State
& Overall \\
\midrule
NICE
& 75.0 $\pm$ 2.5
& 76.5 $\pm$ 2.3
& 86.8 $\pm$ 2.2
& 79.1 $\pm$ 1.8
& 57.3 $\pm$ 2.7
& 51.9 $\pm$ 2.8
& 56.9 $\pm$ 4.1
& 56.0 $\pm$ 5.2
& 56.7 $\pm$ 5.2
& 55.8 $\pm$ 3.5 \\
ATM
& 84.7 $\pm$ 2.0
& 85.4 $\pm$ 2.3
& 90.7 $\pm$ 1.6
& 86.7 $\pm$ 1.8
& 79.5 $\pm$ 3.5
& 76.4 $\pm$ 4.7
& 78.6 $\pm$ 4.5
& \textbf{74.3 $\boldsymbol{\pm}$ 5.8}
& 78.8 $\pm$ 4.6
& 77.7 $\pm$ 4.1 \\
MB2C
& 80.5 $\pm$ 2.6
& 82.0 $\pm$ 2.7
& 88.7 $\pm$ 2.8
& 83.5 $\pm$ 2.5
& 68.2 $\pm$ 3.1
& 64.6 $\pm$ 4.3
& 68.5 $\pm$ 3.3
& 63.5 $\pm$ 3.8
& 68.5 $\pm$ 3.1
& 66.8 $\pm$ 3.1 \\
UBP
& 86.0 $\pm$ 1.7
& 86.1 $\pm$ 1.8
& 95.0 $\pm$ 2.3
& 88.8 $\pm$ 1.8
& 69.3 $\pm$ 2.4
& 66.2 $\pm$ 2.8
& 65.3 $\pm$ 4.0
& 56.8 $\pm$ 3.0
& 68.1 $\pm$ 3.2
& 65.6 $\pm$ 2.1 \\
ATS
& 87.6 $\pm$ 1.7
& 87.8 $\pm$ 1.5
& 94.1 $\pm$ 2.2
& 89.7 $\pm$ 1.7
& 76.3 $\pm$ 2.8
& 69.8 $\pm$ 2.8
& 72.0 $\pm$ 2.7
& 63.2 $\pm$ 2.2
& 72.4 $\pm$ 3.0
& 71.2 $\pm$ 2.1 \\
NeuroBridge
& 88.3 $\pm$ 1.3
& 89.8 $\pm$ 1.3
& 95.7 $\pm$ 1.4
& 91.0 $\pm$ 1.1
& 79.2 $\pm$ 3.6
& 70.2 $\pm$ 2.3
& 71.6 $\pm$ 1.9
& 65.1 $\pm$ 2.8
& 73.5 $\pm$ 1.9
& 72.4 $\pm$ 1.9 \\
HyFI
& 90.1 $\pm$ 1.3
& 92.0 $\pm$ 1.2
& 96.8 $\pm$ 1.7
& 92.8 $\pm$ 1.1
& 81.4 $\pm$ 2.4
& 74.5 $\pm$ 2.3
& 74.9 $\pm$ 2.4
& 69.3 $\pm$ 3.0
& 75.8 $\pm$ 2.7
& 75.7 $\pm$ 1.4 \\
Brain-HIVE
& \textbf{94.2 $\boldsymbol{\pm}$ 1.6}
& \textbf{95.4 $\boldsymbol{\pm}$ 1.7}
& \textbf{97.5 $\boldsymbol{\pm}$ 0.8}
& \textbf{95.6 $\boldsymbol{\pm}$ 1.3}
& \textbf{85.0 $\boldsymbol{\pm}$ 3.4}
& \textbf{81.3 $\boldsymbol{\pm}$ 4.1}
& \textbf{81.3 $\boldsymbol{\pm}$ 4.4}
& 71.8 $\pm$ 4.5
& \textbf{80.6 $\boldsymbol{\pm}$ 4.1}
& \textbf{80.5 $\boldsymbol{\pm}$ 3.6} \\
\bottomrule
\end{tabular}
\caption{Fine-grained and overall 2AFC accuracy for Object Identity and Attribute edits. Attribute subtypes group material with finish, texture with pattern, shape with size, and physical-state changes under State. Best results are shown in bold.}
\label{tab:2afc_subtypes}
\end{table*}

\subsection{Evaluation Protocol}

All evaluations use the source images and edited images that pass quality control.

\paragraph{Standard 200-way retrieval.}
For each EEG query, the conventional protocol ranks the corresponding source image against the other 199 THINGS-EEG2 test images. We report whether the source image is ranked first.

\paragraph{Null-image 200-way retrieval.}
To provide a reference for changes introduced by image edit model, we repeat standard 200-way retrieval after replacing each source candidate with its null-edited counterpart, generated without an intended semantic change.

\paragraph{Edit-pool retrieval.}
For each concept \(c\), the candidate pool is
\[
\mathcal{P}_c=\{I_c^{\mathrm{orig}}\}\cup E_c.
\]
We define \textbf{edit-pool Top-1 accuracy (EP-Top1)} as the proportion of concepts for which the source image is ranked above all valid edited variants derived from the same source image.

\paragraph{Per-family 2AFC accuracy.}
For edit family \(f\), two-alternative forced-choice accuracy measures how often the original image receives a higher similarity score than an edited variant:

\[
A_{\mathrm{2AFC}}^{(f)}
=
\frac{
\sum_{c}
\sum_{I\in E_c^{(f)}}
\mathbf{1}\!\left\{
s(e_c,I_c^{\mathrm{orig}})>s(e_c,I)
\right\}
}{
\sum_{c}|E_c^{(f)}|
}.
\]

where \(\mathbf{1}\{\cdot\}\) denotes the indicator function. Ties are not counted as correct. Overall 2AFC accuracy is computed as a micro-average over all valid edits. EP-Top1 evaluates joint competition within the full edit pool, while 2AFC reports performance separately across edit families.

\section{Experiments}

\subsection{Experimental Setup}
We conduct experiments on THINGS-EEG2~\cite{gifford2022large}, using recordings from ten subjects and the standard subject-dependent split with 200 held-out test concepts. EEG signals are restricted to the first second after stimulus onset, and the 80 repetitions for each test concept are averaged into one query.

We evaluate NICE, MB2C, ATM, UBP, ATS, Brain-HIVE, HyFI, and NeuroBridge using their original model configurations. Each subject-specific model is trained with five random seeds. All methods are evaluated on the same 2,137 quality-controlled edits, which are used only for final testing. We average seeds within each subject and report the mean and standard deviation across subjects. Complete preprocessing, model configurations, training details, and evaluation procedures are provided in the supplementary material.

\begin{figure*}[t]
    \centering
    \includegraphics[width=\textwidth]{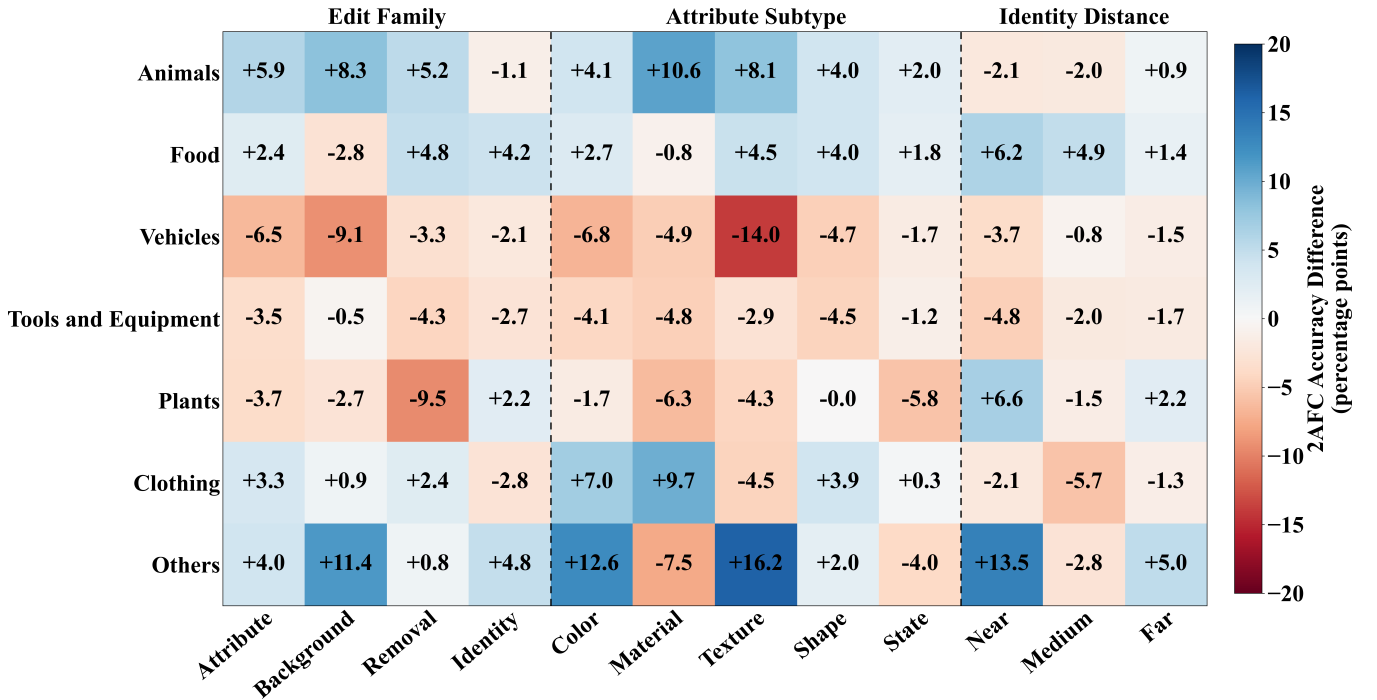}
    \caption{Source-category differences in 2AFC accuracy, averaged across the eight evaluated models. Each cell reports the category-specific 2AFC accuracy minus the all-concept average under the same edit condition, in percentage points. Positive values indicate above-average 2AFC performance within that condition, while negative values indicate below-average 2AFC performance.}
    \label{fig:source_category_analysis}
\end{figure*}

\begin{figure*}[t]
    \centering
    \includegraphics[width=\textwidth]{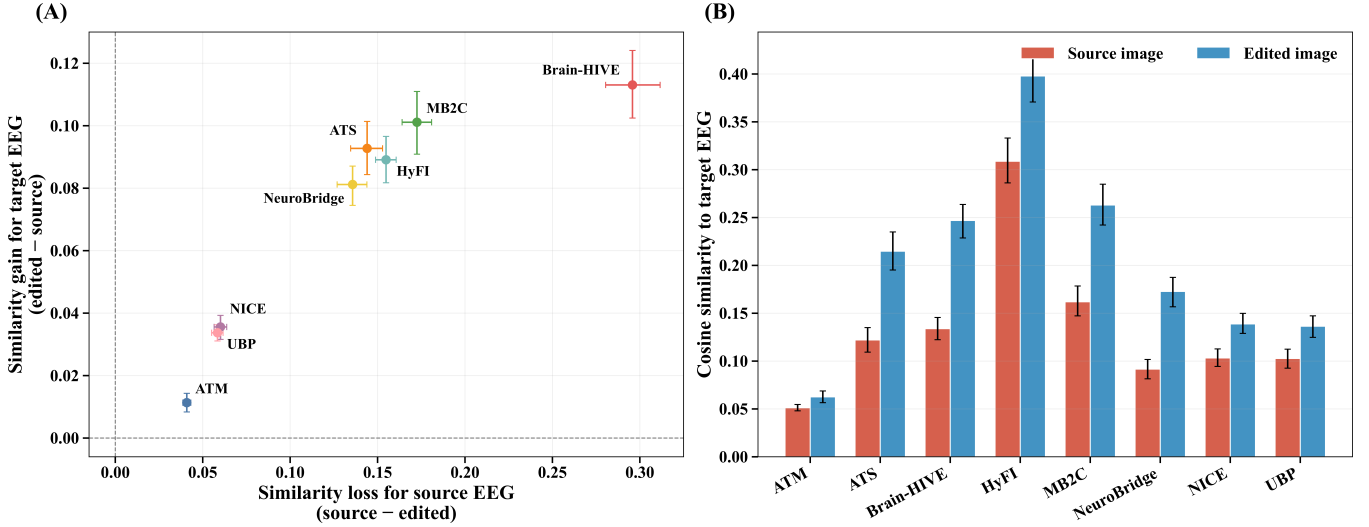}
    \caption{Target-concept alignment under Object Identity Edit. Panel A reports directional similarity changes for each model: source-image similarity minus edited-image similarity for the source-concept EEG on the horizontal axis, and edited-image similarity minus source-image similarity for the target-concept EEG on the vertical axis. Positive values on both axes indicate that the edited image moves away from the source concept and toward the target concept. Panel B compares the similarity scores of the source and edited images to the target-concept EEG. Error bars denote bootstrap 95\% confidence intervals.}
    \label{fig:concept_transfer}
\end{figure*}

\subsection{Semantic Proximity of Edited Images}

Before evaluating EEG retrieval models, we examine how closely the edited images remain anchored to their source images in CLIP space. The colored boxplots in Figure~\ref{fig:semantic_proximity} show the distance between each edited image and its corresponding source image, while the dashed line and shaded region provide a reference for the typical distance between an edited image and images from other test concepts. All distances are computed from OpenCLIP ViT-L/14 embeddings using cosine distance~\cite{radford2021learning,cherti2023reproducible}.

Across all four edit families, edited images generally remain substantially closer to their own source images than to the cross-concept reference. Attribute Edit and Background Edit induce the smallest shifts, reflecting their preservation of the main object and much of the original composition. Object Identity Edit and Object Removal produce larger changes because they replace or remove the primary semantic content, yet their distances still remain below the typical cross-concept level. These results show that EEG-EditBench constructs challenging candidates that preserve substantial source-image content while introducing controlled visual changes.

\subsection{Overall Benchmark Results}

Table~\ref{tab:main_results} summarizes performance under original- and null-image 200-way Top-1 accuracy, EP-Top1, and overall 2AFC. The model rankings differ markedly between the standard and edit-based settings. Brain-HIVE performs best in both retrieval metrics, whereas ATM achieves only moderate 200-way accuracy but remains highly competitive in EP-Top1. By contrast, several models with strong standard retrieval performance lose much of their advantage when the candidate pool is restricted to controlled variants of the same source image. These results show that models successful against unrelated distractors can still confuse the viewed image with closely matched edited variants. EEG-EditBench makes this distinction visible by replacing cross-concept distractors with controlled variants of the viewed image. Replacing the original candidates with null-edited images lowers 200-way accuracy for most models, suggesting that regeneration changes visual representations relevant to retrieval. We examine this effect further in the supplementary material.

A notable pattern appears among methods that incorporate degraded visual representations. UBP, ATS, NeuroBridge, and HyFI all perform strongly in standard 200-way retrieval, yet their advantages shrink substantially in EP-Top1, where ATM surpasses all four despite its lower standard accuracy. Brain-HIVE, which does not rely on degraded visual features, remains strong in both settings. This shared pattern identifies visual degradation as a design factor worth examining under controlled settings.

The overall 2AFC results show a similar separation across models, with Brain-HIVE again achieving the highest accuracy. We next decompose 2AFC performance by edit family and subtype to characterize how model performance varies across edit families and subtypes.

\subsection{Edit-Specific 2AFC Analysis}

The edit-specific results in Table~\ref{tab:main_results} reveal a consistent hierarchy across edit families. Across the eight models, Object Identity Edit achieves 9.0--23.3 percentage points higher accuracy than Attribute Edit, revealing a consistent cross-model advantage in distinguishing changes to object identity. Object Removal also achieves high accuracy, while Background Edit generally lies between Object Identity and Attribute Edit. Within EEG-EditBench, current models therefore capture changes in object identity and presence more reliably than fine-grained attribute changes.

The subtype results in Table~\ref{tab:2afc_subtypes} further refine this picture. Within Object Identity Edit, far replacements are consistently easier to distinguish than near and medium replacements, while the ordering between near and medium varies across models. Semantically distant replacements are therefore more readily distinguished, whereas near and medium replacements remain more challenging. This pattern establishes semantic proximity as a consistent source of difficulty within identity edits. Performance also varies across attribute subtypes: shape or size edits form the most challenging subtype for nearly all models, while color edits are generally the easiest or among the easiest. The variation across attribute subtypes shows that current EEG--image representations preserve different visual properties with different reliability. Together, these results provide a finer-grained view of model behavior than the family-level scores.

\subsection{Source-Category Analysis}

We further examine whether the same edit is equally easy to distinguish across different object categories. We group the 200 test concepts into seven broad source categories and compute category-specific 2AFC accuracy for each edit condition. Figure~\ref{fig:source_category_analysis} reports the difference between each category-specific accuracy and the corresponding all-concept average. The Other category contains only three source images and is not discussed further.

The results show that different categories respond differently to the same attribute edit. Animals perform relatively well on Material and Texture changes, while Clothing shows clearer advantages on Material, Color, and Shape. Attribute sensitivity therefore depends not only on which property is changed, but also on what kind of object is being edited.

The same pattern appears beyond attribute edits. Food performs relatively well on Identity and Removal edits but less well on Background changes, while Vehicles remain below average across several edit conditions. Category differences are largest for near and medium identity replacements and become smaller for far replacements. Source-category differences are therefore most visible when the edited image remains close to the source, while large identity changes are easier to distinguish across categories.

\subsection{Target-Concept Alignment under Object Identity Edit}

The 2AFC results show whether a model distinguishes an object-identity edit from its source image. We further examine whether these edits induce the intended semantic direction in the EEG--image joint space. For edits whose target concept has a corresponding EEG query, we compare the source and edited images with both the source-concept EEG and the target-concept EEG.

Figure~\ref{fig:concept_transfer} shows a consistent target-alignment pattern across all evaluated models. In Panel A, identity editing reduces alignment with the source-concept EEG and increases alignment with the target-concept EEG, indicating a shift from the original concept toward the specified replacement. In Panel B, the edited image also aligns more strongly with the target-concept EEG than the source image. The consistency across models shows that this pattern is shared by different EEG--image alignment architectures. Beyond distinguishing the source and edited images, the analysis characterizes the semantic direction of this distinction in the joint space. Together, the two panels show that object identity edits induce structured, target-directed changes in EEG--image alignment.

\section{Discussion and Conclusion}

The results show that standard retrieval and edit-based evaluation capture different aspects of EEG--image matching. A model that performs well among semantically diverse candidates may still struggle to distinguish the viewed image from closely matched edits. Across the evaluated models, attribute changes are consistently more difficult than changes in object identity or presence, suggesting that current systems preserve coarse semantic information more reliably than fine-grained visual details.

EEG-EditBench is limited by its image-side design. The edited images do not have corresponding EEG recordings, so the benchmark evaluates how a complete EEG--image retrieval system responds to controlled candidate changes rather than how the brain responds to the edited stimuli themselves. In addition, each test concept is represented by a single source image. Future work can address these limitations by introducing greater within-concept diversity, using additional editing models, collecting neural responses to selected edits, and extending the framework to modalities such as fMRI and MEG.

Overall, EEG-EditBench complements standard retrieval with a factor-level view of model behavior under controlled visual changes. The edited variants may also serve as structured hard negatives for developing EEG--image models that retain more precise and interpretable visual information.

\bibliography{aaai2027}

% -----------------------------------------------------------------------------
% Supplementary material
% -----------------------------------------------------------------------------
\appendix
\setcounter{secnumdepth}{2}
\setcounter{section}{0}
\setcounter{figure}{0}
\setcounter{table}{0}
\setcounter{equation}{0}
\renewcommand{\thesection}{S\arabic{section}}
\renewcommand{\thesubsection}{\thesection.\arabic{subsection}}
\renewcommand{\thefigure}{S\arabic{figure}}
\renewcommand{\thetable}{S\arabic{table}}
\renewcommand{\theequation}{S\arabic{equation}}

\section*{Supplementary Material}

\section{Supplementary Overview}

This supplementary material provides the construction, reproduction, and analysis details supporting the main paper. We first describe the benchmark taxonomy, generation pipeline, and quality-control procedure, followed by the preprocessing, model-reproduction, and evaluation protocols. We then report a sensitivity analysis for 2AFC aggregation, validity controls for EEG pairing and image-generation effects, and additional diagnostic results on visual edit magnitude, source categories, target-concept transfer, and representative failures.

\section{Benchmark Construction and Quality Control}

This section describes the benchmark scope and edit taxonomy, the shared construction pipeline, and the human quality-control procedure used to obtain the final dataset.

\subsection{Benchmark Scope and Edit Taxonomy}

EEG-EditBench is constructed from the 200 source images in the THINGS-EEG2 test set~\cite{gifford2022large}, with one image representing each test concept. For every source image, the benchmark introduces controlled candidates that change the identity of the main object, one visible attribute, the surrounding background, or the presence of the object itself. During evaluation, the EEG query elicited by the source image is compared with the source and its edited variants. The quality-controlled edited images are used exclusively for final evaluation.

Before target generation, each source image receives an independent feasibility assessment for the four edit families. Each family is assigned \texttt{accept}, \texttt{borderline}, or \texttt{reject}. An \texttt{accept} decision proceeds to standard target generation, \texttt{borderline} admits conservative targets that remain reliable under the identified image-specific limitation, and \texttt{reject} closes the corresponding generation branch. This family-specific assessment, together with the availability of suitable targets, determines the final source coverage.

Table~\ref{tab:edit_taxonomy} summarizes the controlled change and preserved content for each edit family.

\begin{table*}[t]
\centering
\small
\setlength{\tabcolsep}{5pt}
\begin{tabular}{@{}p{3.0cm}p{4.3cm}p{8.2cm}@{}}
\toprule
Edit family & Controlled change & Preserved content \\
\midrule
Object Identity
& Main object identity
& Composition, viewpoint, background, visual style, and spatial role \\

Attribute
& One visible property of the main object
& Object identity, background, and non-target content \\

Background
& Surrounding scene
& Foreground object, appearance, position, and interaction structure \\

Object Removal
& Complete main subject and reconstruction of the exposed region
& Remaining objects and scene context \\
\bottomrule
\end{tabular}
\caption{Operational definition of the four edit families.}
\label{tab:edit_taxonomy}
\end{table*}

Attribute Edit contains five visible-property groups. Color covers changes in hue, tone, brightness, or saturation; Material/Finish covers perceived substance, reflectance, and surface finish; Texture/Pattern covers markings, grain, print, and other surface structure; Shape/Size covers visible geometric or scale changes that preserve the recognizable object category and spatial role; and State/Condition covers physical states such as age, ripeness, cleanliness, damage, openness, fullness, or wetness.

For Object Identity Edit, replacement targets are drawn from the remaining THINGS-EEG2 test concepts and assigned to three qualitative, scene-conditioned levels. Near targets belong to a similar broad object family or share a closely related function, affordance, or usage context. Medium targets come from a different object family but can occupy a similar role in the scene, interaction, or spatial arrangement. Far targets are semantically distinct concrete objects that remain plausible replacements in the source composition. These assignments jointly consider semantic relatedness and scene compatibility, including approximate scale, spatial role, support relations, and interactions.

Table~\ref{tab:benchmark_composition} reports the resulting subtype counts and source-concept coverage.

\begin{table}[tbp]
\centering
\small
\setlength{\tabcolsep}{4pt}
\begin{tabular}{@{}llcc@{}}
\toprule
Edit family & Subtype & Edits & \shortstack{Source\\concepts} \\
\midrule
Object Identity & Near & 266 & 198 \\
                & Medium & 280 & \\
                & Far & 234 & \\
\addlinespace
Attribute       & Color & 231 & 200 \\
                & Material/Finish & 199 & \\
                & Texture/Pattern & 198 & \\
                & Shape/Size & 156 & \\
                & State/Condition & 195 & \\
\addlinespace
Background      & -- & 199 & 199 \\
Object Removal  & -- & 179 & 179 \\
\midrule
Overall         & -- & 2,137 & 200 \\
\bottomrule
\end{tabular}
\caption{Final benchmark composition by edit family and subtype. Source-concept counts report coverage within each family; the complete benchmark covers all 200 test concepts.}
\label{tab:benchmark_composition}
\end{table}

\subsection{Construction Pipeline}

The construction figure in the main paper summarizes the four-stage workflow used for all source images. Table~\ref{tab:construction_pipeline} provides the input, output, and model used at each stage. The overall workflow is shared across edit families, with family-specific branches for target generation and prompt construction. The complete image-editing settings are reported in Table~\ref{tab:image_editing_configuration}.

\begin{table*}[t]
\centering
\small
\setlength{\tabcolsep}{6pt}
\begin{tabular}{@{}llll@{}}
\toprule
Stage & Input & Output & Model \\
\midrule
Visual Scene Profiling
& Source image and concept label
& Scene profile and editability
& InternVL3.5-38B, multimodal \\

Edit Target Generation
& Scene profile and candidate concepts
& Family-specific edit targets
& InternVL3.5-38B, text-only \\

Edit Prompt Generation
& Scene profile and one target
& Self-contained editing prompt
& InternVL3.5-38B, text-only \\

Image Editing
& Source image and prompt
& One edited image
& FLUX.2-klein-9B \\
\bottomrule
\end{tabular}
\caption{Four-stage benchmark construction pipeline, showing the input, output, and model used at each stage.}
\label{tab:construction_pipeline}
\end{table*}

Edit Target Generation uses four parallel branches. The Attribute branch proposes visible changes to color, material or finish, texture or pattern, shape or size, and state or condition. The Background branch selects a new scene compatible with the foreground subject. The Object Identity branch selects scene-compatible near, medium, and far targets from the remaining test concepts. The Object Removal branch specifies the complete subject to remove and how the exposed region should be reconstructed. Each structured target is then converted into a self-contained editing instruction that states the requested change and the relevant content to preserve, including composition, viewpoint, object placement, visible non-target attributes, and scene relationships.

The first three stages use the local \path{OpenGVLab/InternVL3_5-38B} checkpoint~\cite{wang2025internvl35} with deterministic decoding (\texttt{do\_sample=False}). Each response is validated against a stage-specific schema before being written to an incremental JSONL record. Formatting or constraint violations trigger regeneration with corrective feedback for up to two retries, and completed records are retained when a run is resumed.

The final stage uses \path{black-forest-labs/FLUX.2-klein-9B}~\cite{blackforestlabs2026flux2klein} through \texttt{diffusers.Flux2KleinPipeline}. All four edit families use the same source-image-plus-prompt interface and inference settings. Table~\ref{tab:image_editing_configuration} reports the complete configuration.

\begin{table}[tbp]
\centering
\small
\setlength{\tabcolsep}{6pt}
\begin{tabular}{@{}ll@{}}
\toprule
Setting & Value \\
\midrule
Model & FLUX.2-klein-9B \\
Resolution & 224 \(\times\) 224 \\
Inference steps & 4 \\
Guidance scale & 1.0 \\
Generation seed & 0, reset for each edit \\
Maximum sequence length & 512 \\
Text-encoder output layers & 9, 18, and 27 \\
Precision & bfloat16 or float16 \\
CPU offload & Enabled \\
\bottomrule
\end{tabular}
\caption{Image-editing inference settings.}
\label{tab:image_editing_configuration}
\end{table}

The shared pipeline produced 2,210 candidate edits, which then entered the quality-control procedure.

\subsection{Quality Control and Final Benchmark Composition}

All 2,210 generated edits underwent full human review. Two reviewers independently evaluated every candidate using side-by-side comparisons of the source and edited images together with the corresponding edit instruction. Reviewers assessed whether the intended edit was visibly fulfilled, whether relevant non-target content was preserved, and whether the edited image contained obvious artifacts or implausible visual relations. Each reviewer assigned a binary pass or fail decision without access to the other reviewer's judgment.

Cases receiving different decisions from the two reviewers were independently assessed by a third reviewer, whose judgment determined the final outcome. Only edits receiving a final pass decision were retained. This process retained 2,137 of the 2,210 generated edits.

The final benchmark contains 979 Attribute Edits, 780 Object Identity Edits, 199 Background Edits, and 179 Object Removals. All 200 source concepts remain represented.

\section{Reproduction and Evaluation Details}

This section documents the dataset and EEG preprocessing, the reproduction settings of the eight evaluated models, and the common evaluation and implementation protocol.

\subsection{Dataset and EEG Preprocessing}

We use THINGS-EEG2~\cite{gifford2022large}, which contains EEG recordings from ten subjects acquired with a BrainVision actiCHamp system during a 5~Hz rapid serial visual presentation paradigm. EEG was originally recorded at 1,000~Hz from 63 channels, and each subject completed four recording sessions. All evaluated models use a subject-dependent setting, with a separate model trained and evaluated for each subject. Table~\ref{tab:eeg_dataset_split} summarizes the training and test partitions.

\begin{table}[tbp]
\centering
\small
\setlength{\tabcolsep}{4pt}
\begin{tabular}{@{}lcccc@{}}
\toprule
Partition
& Concepts
& \shortstack{Images per\\concept}
& Images
& \shortstack{Repetitions\\per image} \\
\midrule
Train & 1,654 & 10 & 16,540 & 4 \\
Test  & 200   & 1  & 200    & 80 \\
\bottomrule
\end{tabular}
\caption{THINGS-EEG2 partitions used in this work. Each preprocessed EEG trial contains 63 channels and 250 time samples.}
\label{tab:eeg_dataset_split}
\end{table}

The training and test concepts are disjoint, and the single image associated with each test concept serves as an EEG-EditBench source image.

For preprocessing, trials were first arranged in a fixed 63-channel order, and target and catch trials were removed. EEG was epoched from 0.2~s before to 1.0~s after stimulus onset and baseline-corrected using the prestimulus interval. The signals were then downsampled from 1,000 to 250~Hz, and the 0--1~s post-stimulus interval was retained, yielding 250 samples per channel. Trials were organized by image condition and recording session. For each subject, multivariate noise normalization (MVNN)~\cite{guggenmos2018multivariate} was estimated using only the training partition, and the resulting whitening transform was applied to both training and test EEG. The preprocessed representation of one trial therefore has shape \(63\times250\).

NICE, MB2C, and ATM use all 63 channels. UBP, ATS, Brain-HIVE, HyFI, and NeuroBridge use the following fixed set of 17 occipital and parietal channels: \texttt{P7}, \texttt{P5}, \texttt{P3}, \texttt{P1}, \texttt{Pz}, \texttt{P2}, \texttt{P4}, \texttt{P6}, \texttt{P8}, \texttt{PO7}, \texttt{PO3}, \texttt{POz}, \texttt{PO4}, \texttt{PO8}, \texttt{O1}, \texttt{Oz}, and \texttt{O2}. All methods use the complete 0--1~s post-stimulus window.

At test time, every method averages the 80 repetitions for each image before producing one EEG query embedding. During training, ATM treats the four repetitions of each training image as separate samples, whereas the other seven methods average the four repetitions before model input.

\subsection{Evaluated Models and Training Details}

We evaluate NICE~\cite{song2024decoding}, MB2C~\cite{wei2024mb2c}, ATM~\cite{li2024visual}, UBP~\cite{wu2025bridging}, ATS~\cite{wu2026shrinking}, Brain-HIVE~\cite{zheng2026hierarchical}, HyFI~\cite{jo2026hyfi}, and NeuroBridge~\cite{zhang2026neurobridge} in a subject-dependent setting. Each method is trained independently for ten subjects and five random seeds (0--4), yielding 400 subject--method--seed runs. Each reproduction retains its specified EEG encoder, frozen visual representation, and training objective. The resulting EEG and image embeddings are exported to a common evaluator. EEG-EditBench edited images are used only in the final evaluation and do not contribute to training or checkpoint selection. Table~\ref{tab:model_overview} summarizes the model-specific representations and objectives.

\begin{table*}[t]
\centering
\small
\setlength{\tabcolsep}{7pt}
\begin{tabular}{@{}llllc@{}}
\toprule
Method & EEG encoder & Visual target & Objective & Dim. \\
\midrule
NICE
& Temporal--spatial CNN
& CLIP ViT-L/14
& Symmetric contrastive
& 768 \\

MB2C
& NICE-style CNN
& CLIP ViT-L/14
& Contrastive, cycle, and adversarial
& 768 \\

ATM
& Transformer + ShallowNet
& OpenCLIP ViT-H/14 image and text
& Image--text alignment
& 1,024 \\

UBP
& EEGProjectLayer
& RN50 multi-level blur
& Uncertainty-aware contrastive
& 1,024 \\

ATS
& EEGProjectLayer
& RN50 adaptive blur
& Adaptive-teaching contrastive
& 768 \\

Brain-HIVE
& Brain MLP
& SynCLR, CLIP, and SDXL-VAE
& Hierarchical visual alignment
& 1,024 \\

HyFI
& EEGProjectLayer
& RN50 semantic and perceptual blur
& Hyperbolic contrastive
& 1,024 \\

NeuroBridge
& EEGProject
& Augmented-fused RN50
& Symmetric contrastive
& 512 \\
\bottomrule
\end{tabular}
\caption{Overview of the eight reproduced EEG--image models. Dim. denotes the shared EEG and image embedding dimension.}
\label{tab:model_overview}
\end{table*}

Table~\ref{tab:training_protocols} reports the principal optimization settings. Our reproductions are based on the protocols described in the original works and official implementations.

\begin{table}[t]
\centering
\footnotesize
\setlength{\tabcolsep}{3pt}
\begin{tabularx}{\columnwidth}{@{}lrr>{\raggedright\arraybackslash}X@{}}
\toprule
Method & Epochs & Batch & Optimization \\
\midrule
NICE
& 200
& 1,000
& Adam, \(2\times10^{-4}\) \\

MB2C
& \(\leq\)1,000
& 2,048
& Adam / RMSprop \\

ATM
& 80
& 1,024
& AdamW, \(3\times10^{-4}\) \\

UBP
& 50
& 1,024
& AdamW, \(1\times10^{-4}\) \\

ATS
& 150
& 1,024
& AdamW, \(1\times10^{-4}\); StepLR \\

Brain-HIVE
& 25
& 1,024
& AdamW, \(5\times10^{-4}\); cosine \\

HyFI
& \(\leq\)50
& 1,024
& AdamW, \(3\times10^{-4}\) \\

NeuroBridge
& 50
& 1,024
& AdamW, \(1\times10^{-4}\) \\
\bottomrule
\end{tabularx}
\caption{Principal training settings for the reproduced models. Batch denotes the effective batch size of one subject-specific run.}
\label{tab:training_protocols}
\end{table}

\paragraph{CLIP-based image and image--text alignment.}
NICE randomly divides its training data into a 740-example validation subset and the remaining training examples. MB2C augments symmetric InfoNCE with bidirectional cycle consistency, bidirectional adversarial learning, auxiliary classification, and mixup. Its principal settings are a cycle weight of 10, a mixup ratio of 0.75, a gradient-penalty weight of 10, and an auxiliary-classification weight of 1. ATM combines image and text alignment losses with weights 0.99 and 0.01, respectively, using text targets generated from the template \texttt{This picture is \{concept\}}.

\paragraph{Blur-based visual supervision.}
UBP constructs three FoveaBlur representations with kernel sizes 45, 51, and 57 and dynamically selects among them according to the positive EEG--image similarity. Both source and edited images use the medium, kernel-51 representation during evaluation. ATS follows the same three-level adaptive-teaching principle and uses a ShrinkAdapter with a bottleneck ratio of 0.25 to map the 1,024-dimensional RN50 features to 768 dimensions. HyFI combines a semantic FoveaBlur branch with kernel size 51 and a perceptual blur branch with kernel size 31, maps EEG and both visual branches to Lorentz space, and learns a geodesic interpolation of the semantic and perceptual targets.

\paragraph{Hierarchical and augmented visual supervision.}
Brain-HIVE uses 768-dimensional SynCLR ViT-B/16 features, 512-dimensional CLIP ViT-B/32 features, and 1,024-dimensional SDXL-VAE latents. These branches are projected into a shared 1,024-dimensional space and fused during Stage-2 training, which uses bfloat16 and a cosine learning-rate schedule without warmup. NeuroBridge constructs its visual target by fusing deterministic OpenCLIP RN50 features from Gaussian blur, Gaussian noise, low-resolution, and mosaic transformations. The formal configuration uses this visual branch without an additional text target. During training, random smooth EEG augmentation with kernel size 5 is applied with probability 0.3.

\subsection{Evaluation Protocol and Implementation}

\paragraph{Evaluation scores.}
For every subject--method--seed run, the evaluator receives one EEG query representation for each of the 200 test concepts, together with image representations for the 200 source images and all 2,137 retained edits. Following the original evaluation protocol of each method, we use its native EEG--image score for retrieval and pairwise comparison. All reported metrics are computed within each model run.

\paragraph{Standard 200-way retrieval.}
For each EEG query, the standard protocol ranks all 200 source images by the corresponding method-native score. Top-1 and Top-5 accuracy measure whether the corresponding source image appears among the first one or five candidates. Their random-ranking references are \(1/200=0.5\%\) and \(5/200=2.5\%\), respectively.

\paragraph{Edit-pool retrieval.}
For each test concept \(c\), the complete candidate pool contains its source image and all retained edits derived from that source; let \(\mathcal{P}_c\) denote this pool. Edit-pool Top-1 accuracy (EP-Top1) requires the source image to rank above every same-source edit:
\begin{equation}
A_{\mathrm{EP}}
=
\frac{1}{200}
\sum_{c=1}^{200}
\mathbf{1}
\!\left[
s(e_c,v_c^{\mathrm{source}})
>
\max_i s(e_c,v_{c,i}^{\mathrm{edit}})
\right].
\label{eq:ep_top1}
\end{equation}
Each concept contributes equally. The 200 pools contain 5--14 edited candidates and one source image, giving total pool sizes of 6--15 and a mean size of 11.685. Because pool sizes vary, the random-ranking reference is computed concept by concept:
\begin{equation}
A_{\mathrm{EP}}^{\mathrm{chance}}
=
\frac{1}{200}
\sum_{c=1}^{200}
\frac{1}{|\mathcal{P}_c|}
=0.08722\approx8.72\%.
\label{eq:ep_chance}
\end{equation}

\paragraph{Per-edit 2AFC accuracy.}
Each retained edit defines a two-alternative comparison between that edit and its corresponding source image. For edit family \(f\),
\begin{equation}
A_{\mathrm{2AFC}}^{(f)}
=
\frac{
\displaystyle\sum_c\sum_{i\in E_c^{(f)}}
\mathbf{1}
\!\left[
s(e_c,v_c^{\mathrm{source}})
>
s(e_c,v_{c,i}^{\mathrm{edit}})
\right]
}{
\displaystyle\sum_c |E_c^{(f)}|
}.
\label{eq:2afc_family}
\end{equation}
The source must receive a strictly higher similarity; a tie is not counted as correct. Overall 2AFC is the micro-average over all 2,137 edits. The same definition is applied separately to the 979 Attribute, 780 Object Identity, 199 Background, and 179 Object Removal edits. The random-ranking reference is 50\%. Table~\ref{tab:evaluation_protocol_summary} summarizes the three protocols.

\begin{table}[tbp]
\centering
\small
\setlength{\tabcolsep}{4pt}
\begin{tabular}{@{}lcc@{}}
\toprule
Protocol & Candidates & Chance \\
\midrule
Standard Top-1/5
& 200 sources
& 0.5\% / 2.5\% \\

EP-Top1
& Source + edits
& 8.72\% \\

2AFC
& Source vs.\ one edit
& 50\% \\
\bottomrule
\end{tabular}
\caption{Summary of the three evaluation protocols and their random-ranking references. EP-Top1 compares each source with all same-source edits. The protocols contain 200 concept-level queries, 200 concept-level pools, and 2,137 edit-level pairs, respectively.}
\label{tab:evaluation_protocol_summary}
\end{table}

\paragraph{Aggregation across seeds and subjects.}
Let \(m_{s,r}\) denote a metric from subject \(s\) and training seed \(r\). We first average the five seeds within each subject,
\begin{equation}
\bar{m}_s=\frac{1}{5}\sum_{r=0}^{4}m_{s,r},
\label{eq:seed_aggregation}
\end{equation}
and then report the mean and population standard deviation across the ten subject-level values:
\begin{equation}
\mu=\frac{1}{10}\sum_{s=1}^{10}\bar{m}_s,
\qquad
\sigma=
\sqrt{
\frac{1}{10}
\sum_{s=1}^{10}(\bar{m}_s-\mu)^2
}.
\label{eq:subject_aggregation}
\end{equation}
The reported uncertainty thus reflects variation across subjects after seed averaging, with each subject serving as the unit of analysis.

\paragraph{Implementation environment.}
Table~\ref{tab:implementation_environment} lists the principal software and hardware used for model reproduction and evaluation. Formal runners pass seeds explicitly to Python, NumPy, and PyTorch and record the method, subject, and seed in each run's metadata.

\begin{table}[tbp]
\centering
\small
\setlength{\tabcolsep}{5pt}
\begin{tabular}{@{}ll@{}}
\toprule
Component & Version or configuration \\
\midrule
Python & 3.9.23 \\
PyTorch & 2.5.0, CUDA 11.8 build \\
NumPy & 1.26.4 \\
SciPy & 1.13.1 \\
scikit-learn & 1.6.1 \\
PyTorch Lightning & 2.6.0 \\
Transformers & 4.57.6 \\
OpenCLIP & 3.2.0 \\
MNE & 1.8.0 \\
GPU & NVIDIA RTX 3090, 24 GB \\
\bottomrule
\end{tabular}
\caption{Principal implementation environment.}
\label{tab:implementation_environment}
\end{table}

\paragraph{Computing and caching.}
The server provides eight RTX 3090 GPUs. Each MB2C run uses two GPUs with data parallelism, whereas each run of the other seven models uses one GPU. Only subject-independent features from frozen image encoders are cached; EEG models and EEG query embeddings are never reused across runs. Brain-HIVE visual caches are isolated by seed because its VAE branch includes stochastic sampling. Before reuse, cache metadata verifies the encoder checkpoint, preprocessing configuration, manifest, input order, file identities, and feature shapes.

\begin{figure}[t]
\centering
\includegraphics[width=\columnwidth]{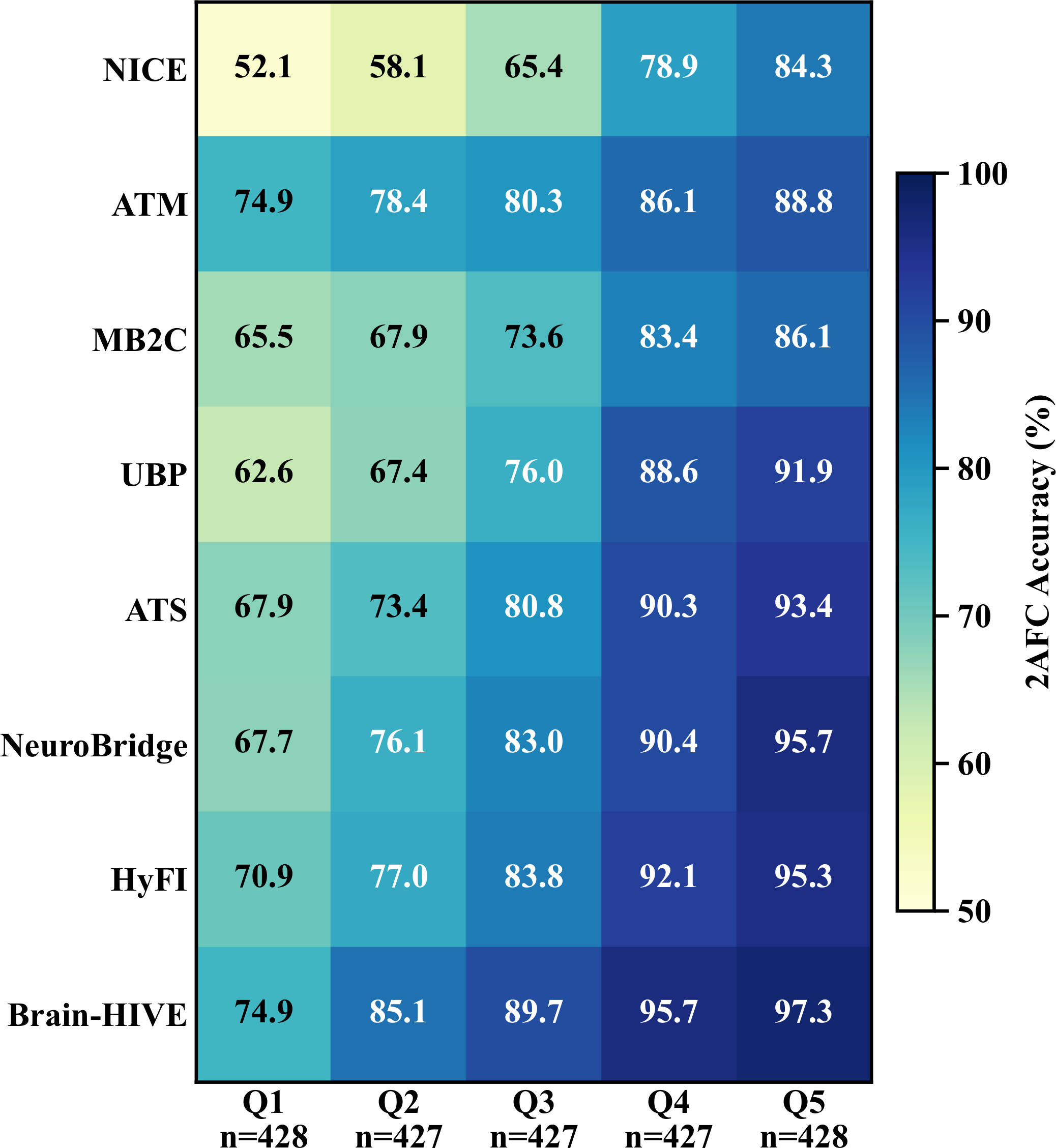}
\caption{2AFC accuracy across source--edit visual-distance quintiles, from the smallest changes in Q1 to the largest in Q5. The OpenCLIP cosine-distance intervals are 0.013--0.099, 0.099--0.149, 0.149--0.236, 0.236--0.310, and 0.310--0.575; edit counts are shown below the columns.}
\label{fig:visual_distance_2afc}
\end{figure}

\begin{figure*}[t]
\centering
\includegraphics[width=\textwidth]{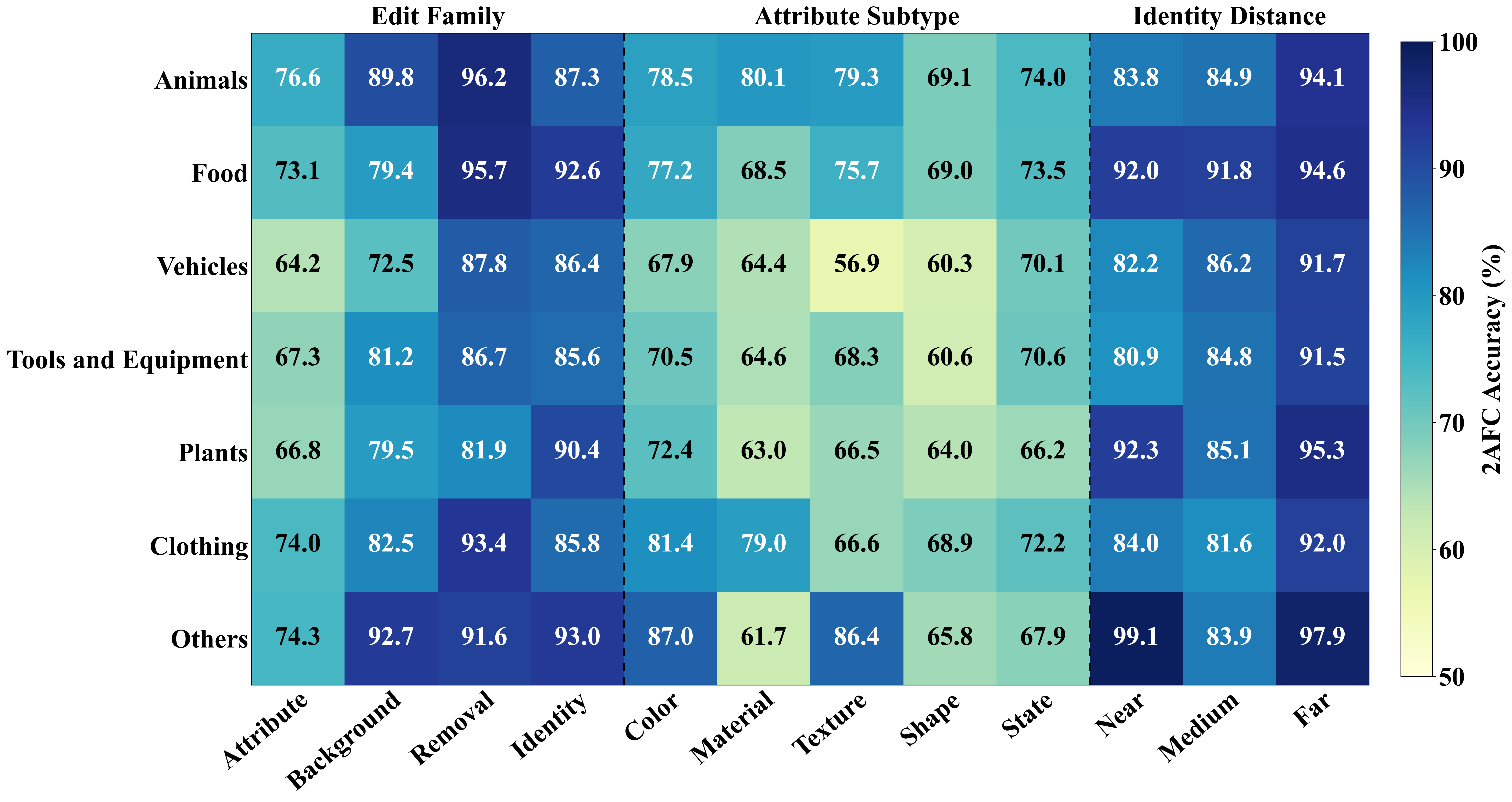}
\caption{Absolute source-category 2AFC accuracy across the four edit families, five Attribute subtypes, and three Object Identity distance levels. Rows denote the seven source categories.}
\label{fig:absolute_source_category_2afc}
\end{figure*}

\begin{figure*}[t]
\centering
\includegraphics[width=\textwidth]{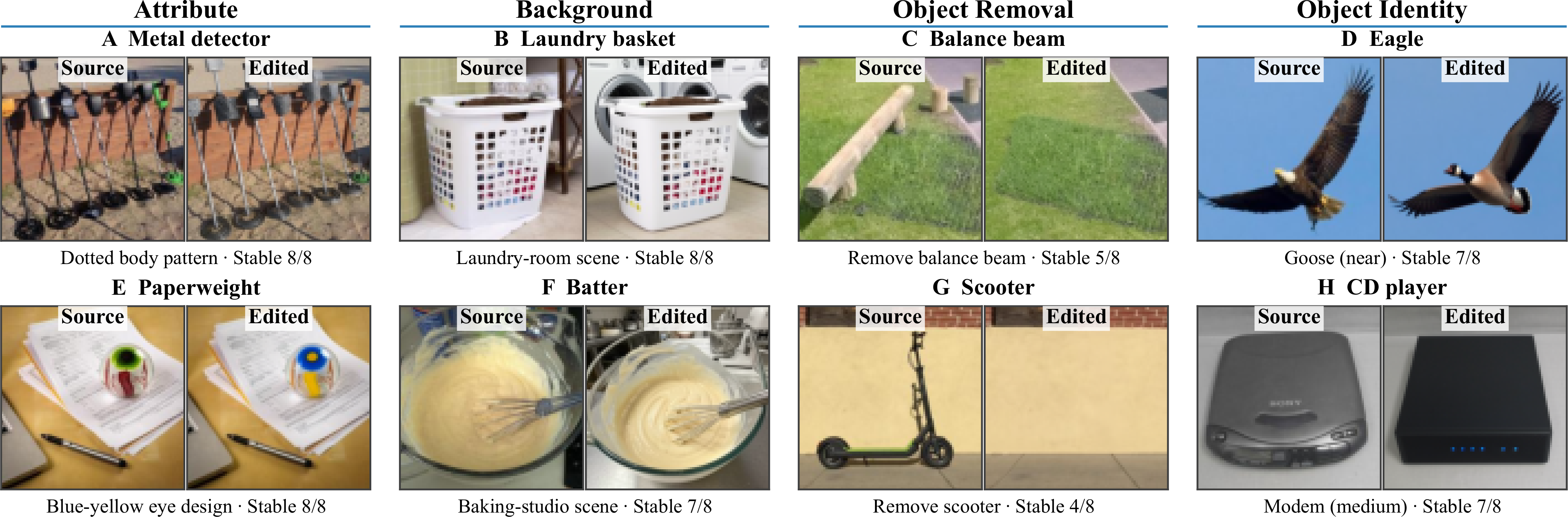}
\caption{Representative stable failure cases across the four edit families. Columns denote edit families, and the upper and lower panels show two cases with different source images. Each panel reports the number of models satisfying the stable-failure criterion.}
\label{fig:representative_stable_failures}
\end{figure*}

\begin{figure}[t]
\centering
\includegraphics[width=\columnwidth]{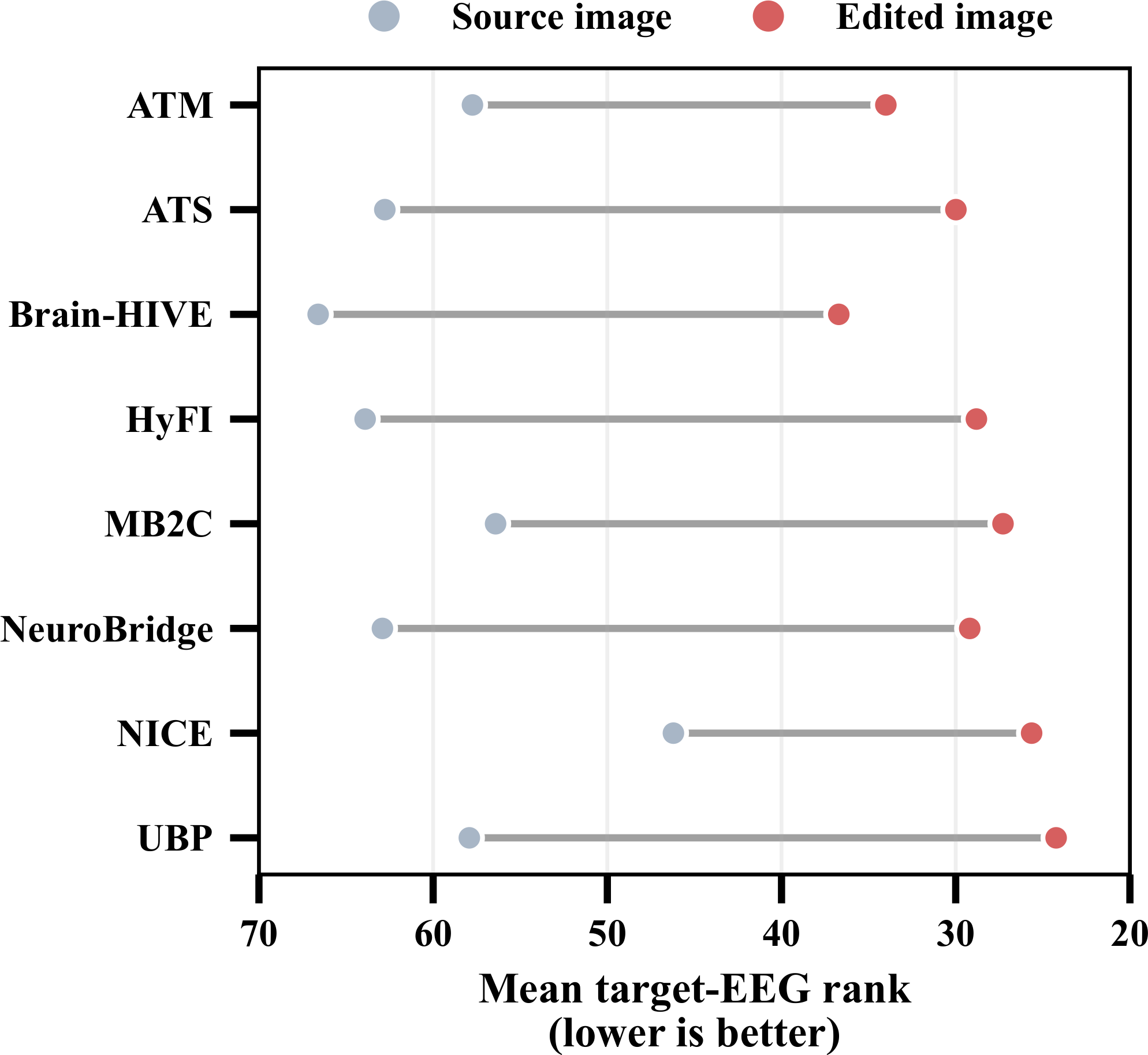}
\caption{Mean rank of the specified target-concept EEG among all 200 EEG concepts for source and edited images. Lower rank indicates stronger target-concept alignment.}
\label{fig:target_concept_rank_shift}
\end{figure}

\section{Evaluation Sensitivity}

We examine whether the main 2AFC patterns depend on the weighting assigned to concepts and edit families.

\subsection{Alternative 2AFC Aggregation}

The overall 2AFC result in the main paper gives equal weight to every retained edit, so concepts and edit families with more examples contribute more to the final score. We compare this edit-balanced result with a concept-and-family-balanced alternative that gives equal total weight to each edit family and equal weight to source concepts within each family, using the same 2,137 edits and strict source-over-edit decision rule. Table~\ref{tab:alternative_2afc_aggregations} reports the comparison.

The balanced aggregation produces modest changes in absolute accuracy while preserving the main model and edit-family patterns. The maximum model-rank change is one position.

\section{Validity Controls}

We examine two factors that directly affect the interpretation of edit-based evaluation: dependence on the correct EEG--image correspondence and sensitivity to characteristics introduced by image generation.

\subsection{Dependence on EEG--Image Pairing}

We first examine how strongly source--edit discrimination depends on the correct EEG--image correspondence. We compare each source--edit pair under the correctly matched EEG query and under all 199 non-matching EEG queries. The image candidates remain fixed, and the same mismatch schedule is used for every model, subject, and training run. Correctly paired and mismatched-query 2AFC use the same strict source-over-edit decision rule. We define the pairing gain as
\[
\Delta_{\mathrm{pair}}=A_{\mathrm{correct}}-A_{\mathrm{mismatch}}.
\]
Metrics are averaged across seeds within subject, and confidence intervals are obtained by bootstrapping the ten subject-level values. Table~\ref{tab:mismatched_eeg_overall} reports the overall comparison.

Correct pairing improves 2AFC for all eight models, with gains of 13.89--35.14 percentage points. The confidence intervals are above zero in every case, showing that source--edit discrimination consistently benefits from the correct EEG--image correspondence. Table~\ref{tab:mismatched_eeg_by_family} provides the corresponding breakdown by edit family.

The family-level results show the same general pattern, with positive gains in nearly all model--family combinations. NICE on Attribute edits is the only case whose confidence interval includes zero. Mismatched-query performance also varies across systems, indicating that model-specific image preferences coexist with the correspondence-dependent effect.

\subsection{Sensitivity to Image-Generation Effects}

We use one null-edited image for each test concept. The null-edit instruction asks the editor to reproduce the source image while preserving its depicted objects, visible attributes, composition, and background, without introducing an intended semantic change. Null edits use the same image-editing model, resolution, inference settings, and generation seed as the semantic edits, and are reviewed side by side with their source images to exclude obvious semantic changes or generation failures. We compare each original image with its null-edited counterpart, and then compare the null-edited image with each semantic edit in EEG-EditBench.

The first comparison provides a reference for changes associated with regeneration. In the second, both candidates have passed through the editing pipeline, and we evaluate them with correctly paired and mismatched EEG queries. Pairing gain is the difference between these two EEG conditions. All comparisons use the same edit-level weighting over the 2,137 retained semantic edits, and confidence intervals are computed over subjects. Table~\ref{tab:generated_candidate_2afc} reports the results.

Original-over-null 2AFC is above the 50\% reference for seven models, while NICE remains close to balance, showing that most systems are sensitive to the image-side changes introduced by regeneration. When both candidates are generated images, correctly paired EEG outperforms mismatched EEG for every model, with pairing gains of 16.23--29.29 percentage points.

\section{Additional Diagnostic Analyses}

We further characterize benchmark behavior through visual edit magnitude, source-category patterns, target-concept alignment, and representative stable failures.

\subsection{Evaluation Behavior across Visual Edit Magnitudes}

We next examine how visual edit magnitude shapes 2AFC difficulty and edit-family differences.

\paragraph{2AFC across visual distances.}
We measure edit magnitude using cosine distance between the source and edited images in a common OpenCLIP ViT-L/14 feature space~\cite{radford2021learning,cherti2023reproducible} that is independent of the eight evaluated EEG--image systems. For edit $i$,
\begin{equation}
d_i
=
1-
\frac{
\mathbf{v}^{\mathrm{source}}_i \cdot \mathbf{v}^{\mathrm{edit}}_i
}{
\lVert \mathbf{v}^{\mathrm{source}}_i \rVert_2
\lVert \mathbf{v}^{\mathrm{edit}}_i \rVert_2
}.
\label{eq:visual_edit_distance}
\end{equation}

The 2,137 edits are divided into five shared distance quintiles, which are reused for every model, subject, and training seed. Figure~\ref{fig:visual_distance_2afc} reports 2AFC accuracy in each quintile.

All eight models show non-decreasing 2AFC accuracy from Q1 to Q5, with differences of 13.89--32.15 percentage points between the two extremes. We also correlate visual distance with the source-minus-edit similarity margin within each run. The model-mean Spearman correlations are positive for all eight systems and range from 0.417 to 0.608. Larger visual changes are therefore consistently associated with easier source--edit discrimination.

\paragraph{Distance-balanced Attribute comparisons.}
We next test whether the lower 2AFC accuracy of Attribute edits is primarily associated with their visual-distance distribution. Each comparison is restricted to source concepts represented in both families. Distances are divided into intervals of width 0.05, retaining intervals with at least 20 edits from each family, and the two families receive equal total weight within every retained interval. The unbalanced and distance-balanced estimates use the same eligible edits and differ only in weighting. Table~\ref{tab:distance_balanced_family_gaps} reports comparisons of Attribute with Background, Object Removal, and Object Identity.

Distance balancing reduces several absolute differences while preserving the overall ordering across edit families. Across all eight systems, Attribute Edit remains the most challenging, followed by Object Removal and Object Identity.

\subsection{Absolute Source-Category Results}

The main paper reports source-category performance relative to the all-concept average under each edit condition. Figure~\ref{fig:absolute_source_category_2afc} provides the corresponding absolute 2AFC accuracies. Results are averaged across training seeds within subject, then across subjects within model, and finally across the eight evaluated models.

Vehicles--Texture is the most difficult displayed combination at 56.9\%, whereas far Object Identity edits reach 91.5\%--97.9\% across source categories. Category differences remain visible for near and medium replacements but narrow once the identity change is far from the source concept. The Other category contains only three source images and is included for completeness rather than interpreted individually.

\subsection{Target-Concept Rank Shift under Object Identity Edits}

The target-concept analysis in the main paper measures whether Object Identity edits move image representations away from the source concept and toward the specified target concept. We complement this analysis by ranking the specified target-concept EEG among all 200 EEG concepts for both the source and edited images. Lower rank indicates stronger target-concept alignment. Figure~\ref{fig:target_concept_rank_shift} reports the results across the complete Object Identity Edit set.
The specified target concept moves closer to the front of the complete ranking after editing for all eight models. This candidate-set view complements the similarity analysis in the main paper by showing that Object Identity edits consistently improve the relative position of their intended target concepts.

\subsection{Representative Stable Failure Cases}

We finally examine edits that repeatedly cause one or more systems to prefer the edited image over the viewed source image. For each model--edit pair, a stable failure requires at least 26 failures among the 50 subject--seed comparisons, with at least three seeds each producing failures for at least six of the ten subjects.

Within each edit family, qualifying edits are ranked by cross-model coverage and total failures, and two cases with different source images are selected. Figure~\ref{fig:representative_stable_failures} shows the resulting examples.

Stable failures occur in all four edit families. The two selected Attribute cases affect all eight models and show the broadest cross-model coverage, while the remaining examples demonstrate persistent failures under changes to scene context, object presence, and object identity. These cases provide concrete illustrations of the model behavior summarized by the benchmark-level results.

\section{Generative AI Use Disclosure}

Generative AI tools were used to assist with language editing and drafting during manuscript preparation. The authors reviewed and revised all AI-assisted content and take full responsibility for the manuscript, including its claims, analyses, figures, and references.

\begin{table}[tbp]
\centering
\small
\setlength{\tabcolsep}{7pt}
\begin{tabular}{@{}lcc@{}}
\toprule
Model
& Edit-balanced
& Concept-and-family-balanced \\
\midrule
NICE
& 67.7 \(\pm\) 2.3
& 71.4 \(\pm\) 2.7 \\

ATM
& 81.7 \(\pm\) 2.7
& 82.0 \(\pm\) 2.3 \\

MB2C
& 75.3 \(\pm\) 2.5
& 77.5 \(\pm\) 2.7 \\

UBP
& 77.3 \(\pm\) 1.8
& 80.6 \(\pm\) 1.9 \\

ATS
& 81.1 \(\pm\) 1.6
& 84.9 \(\pm\) 1.6 \\

NeuroBridge
& 82.6 \(\pm\) 1.1
& 86.8 \(\pm\) 1.0 \\

HyFI
& 84.9 \(\pm\) 1.1
& 88.6 \(\pm\) 1.1 \\

Brain-HIVE
& \textbf{88.6 \(\pm\) 2.2}
& \textbf{91.6 \(\pm\) 1.8} \\
\bottomrule
\end{tabular}
\caption{Overall 2AFC accuracy under the formal edit-balanced aggregation and a concept-and-family-balanced alternative. Values are percentages reported as mean \(\pm\) population standard deviation across subjects after within-subject aggregation. The best result in each column is shown in bold.}
\label{tab:alternative_2afc_aggregations}
\end{table}

\begin{table*}[t]
\centering
\small
\setlength{\tabcolsep}{5pt}
\begin{tabular}{@{}lccc@{}}
\toprule
Model
& Correctly paired 2AFC
& Mismatched-query 2AFC
& Pairing gain \\
\midrule
NICE
& 67.73 [66.34, 69.21]
& 53.84 [52.07, 55.70]
& 13.89 [13.19, 14.50] \\
ATM
& 81.71 [80.05, 83.38]
& 59.82 [57.57, 62.40]
& 21.89 [20.56, 23.24] \\
MB2C
& 75.28 [73.69, 76.79]
& 48.52 [47.20, 49.86]
& 26.75 [25.80, 27.74] \\
UBP
& 77.31 [76.15, 78.42]
& 49.92 [49.37, 50.45]
& 27.39 [26.48, 28.31] \\
ATS
& 81.15 [80.16, 82.20]
& 46.51 [45.99, 47.08]
& 34.64 [33.63, 35.70] \\
NeuroBridge
& 82.59 [81.90, 83.26]
& 49.95 [49.65, 50.25]
& 32.64 [32.10, 33.20] \\
HyFI
& 84.94 [84.27, 85.59]
& 51.60 [51.27, 51.94]
& 33.34 [32.60, 34.02] \\
Brain-HIVE
& 88.57 [87.19, 89.92]
& 53.43 [52.55, 54.21]
& 35.14 [34.21, 36.10] \\
\bottomrule
\end{tabular}
\caption{Overall 2AFC under correctly paired and mismatched EEG queries. Pairing gain is their difference in percentage points. Brackets report subject-level 95\% bootstrap confidence intervals.}
\label{tab:mismatched_eeg_overall}
\end{table*}

\begin{table*}[t]
\centering
\small
\setlength{\tabcolsep}{3pt}
\begin{tabular}{@{}lcccc@{}}
\toprule
Model
& \shortstack{Original over null\\Correct EEG}
& \shortstack{Null over semantic edit\\Correct EEG}
& \shortstack{Null over semantic edit\\Mismatched EEG}
& Pairing gain \\
\midrule
NICE
& 49.83 [47.25, 52.50]
& 67.57 [66.80, 68.37]
& 50.12 [49.30, 50.97]
& 17.45 [16.87, 18.04] \\
ATM
& 67.99 [64.24, 71.75]
& 73.13 [71.52, 74.94]
& 56.90 [55.93, 57.99]
& 16.23 [15.44, 17.00] \\
MB2C
& 67.75 [64.67, 70.76]
& 67.34 [66.60, 68.01]
& 49.52 [49.05, 49.94]
& 17.82 [17.22, 18.30] \\
UBP
& 56.43 [55.04, 57.83]
& 73.56 [72.58, 74.47]
& 47.51 [47.00, 48.03]
& 26.05 [25.07, 27.05] \\
ATS
& 61.54 [59.06, 63.80]
& 75.87 [74.74, 77.01]
& 53.53 [53.13, 53.92]
& 22.33 [21.32, 23.34] \\
NeuroBridge
& 58.66 [56.84, 60.56]
& 80.23 [79.47, 81.07]
& 51.30 [50.97, 51.62]
& 28.93 [28.23, 29.66] \\
HyFI
& 66.32 [64.12, 68.51]
& 78.83 [77.85, 79.76]
& 49.54 [48.87, 50.26]
& 29.29 [28.41, 30.17] \\
Brain-HIVE
& 78.10 [74.29, 81.90]
& 74.38 [73.28, 75.45]
& 48.08 [47.31, 48.83]
& 26.30 [25.53, 27.11] \\
\bottomrule
\end{tabular}
\caption{2AFC comparisons involving null-edited and semantic-edit candidates. Pairing gain is the difference between correctly paired and mismatched EEG conditions, in percentage points. Brackets report subject-level 95\% bootstrap confidence intervals.}
\label{tab:generated_candidate_2afc}
\end{table*}

\begin{table*}[t]
\centering
\small
\setlength{\tabcolsep}{5pt}
\begin{tabular}{@{}lcccc@{}}
\toprule
Model & Attribute & Background & Object Removal & Object Identity \\
\midrule
NICE
& 0.59 [-0.50, 1.69]
& 13.10 [11.77, 14.56]
& 29.60 [27.98, 31.16]
& 27.17 [26.27, 27.95] \\
ATM
& 16.08 [15.00, 17.20]
& 21.26 [19.61, 22.92]
& 18.28 [16.22, 20.47]
& 30.17 [28.41, 31.87] \\
MB2C
& 18.41 [16.84, 19.93]
& 28.33 [27.05, 29.37]
& 29.47 [27.69, 31.40]
& 36.20 [35.21, 37.20] \\
UBP
& 17.86 [16.95, 18.79]
& 23.84 [22.87, 24.74]
& 37.87 [35.92, 39.76]
& 37.86 [36.64, 39.11] \\
ATS
& 24.35 [23.21, 25.63]
& 43.35 [42.44, 44.23]
& 29.50 [27.99, 30.85]
& 46.51 [45.06, 47.86] \\
NeuroBridge
& 21.99 [21.17, 22.87]
& 42.12 [41.41, 42.93]
& 24.86 [23.71, 26.13]
& 45.38 [44.44, 46.22] \\
HyFI
& 25.36 [24.56, 26.03]
& 33.57 [32.39, 34.98]
& 26.98 [25.17, 28.97]
& 44.75 [43.86, 45.65] \\
Brain-HIVE
& 27.77 [26.09, 29.44]
& 36.36 [35.19, 37.39]
& 31.50 [29.72, 33.08]
& 44.91 [44.15, 45.67] \\
\bottomrule
\end{tabular}
\caption{Pairing gain by edit family, in percentage points. Brackets report subject-level 95\% bootstrap confidence intervals.}
\label{tab:mismatched_eeg_by_family}
\end{table*}

\begin{table*}[t]
\centering
\small
\setlength{\tabcolsep}{2.8pt}
\begin{tabular}{@{}lcccccc@{}}
\toprule
& \multicolumn{2}{c}{Background \(\boldsymbol{-}\) Attribute}
& \multicolumn{2}{c}{Object Removal \(\boldsymbol{-}\) Attribute}
& \multicolumn{2}{c}{Object Identity \(\boldsymbol{-}\) Attribute} \\
\cmidrule(lr){2-3}\cmidrule(lr){4-5}\cmidrule(lr){6-7}
Model & Unbalanced & Distance-balanced & Unbalanced & Distance-balanced & Unbalanced & Distance-balanced \\
\midrule
NICE        & +10.22 \(\pm\) 5.74 & +8.14 \(\pm\) 5.39  & +16.72 \(\pm\) 4.09 & +14.13 \(\pm\) 5.13 & +14.66 \(\pm\) 4.32 & +8.78 \(\pm\) 4.37 \\
ATM         & \(-\)0.84 \(\pm\) 5.23  & \(-\)1.62 \(\pm\) 4.97  & +7.65 \(\pm\) 4.51  & +6.78 \(\pm\) 3.91  & +3.94 \(\pm\) 2.37  & +1.73 \(\pm\) 3.11 \\
MB2C        & +9.86 \(\pm\) 3.28  & +9.07 \(\pm\) 3.62  & +7.15 \(\pm\) 4.39  & +5.01 \(\pm\) 4.42  & +9.80 \(\pm\) 4.47  & +5.00 \(\pm\) 4.44 \\
UBP         & +13.89 \(\pm\) 2.42 & +12.63 \(\pm\) 2.67 & +17.32 \(\pm\) 3.97 & +13.51 \(\pm\) 4.14 & +15.52 \(\pm\) 2.07 & +11.66 \(\pm\) 3.32 \\
ATS         & +12.77 \(\pm\) 3.59 & +11.05 \(\pm\) 3.75 & +15.81 \(\pm\) 3.39 & +12.71 \(\pm\) 3.66 & +10.67 \(\pm\) 2.09 & +7.36 \(\pm\) 2.52 \\
NeuroBridge & +13.25 \(\pm\) 3.08 & +10.80 \(\pm\) 3.03 & +16.48 \(\pm\) 2.91 & +15.47 \(\pm\) 2.89 & +10.13 \(\pm\) 2.60 & +7.04 \(\pm\) 2.51 \\
HyFI        & +12.06 \(\pm\) 2.22 & +10.09 \(\pm\) 2.20 & +17.04 \(\pm\) 3.56 & +15.00 \(\pm\) 3.52 & +10.71 \(\pm\) 2.33 & +8.67 \(\pm\) 3.21 \\
Brain-HIVE  & +10.36 \(\pm\) 3.34 & +8.41 \(\pm\) 3.57  & +8.37 \(\pm\) 5.76  & +7.15 \(\pm\) 5.42  & +9.02 \(\pm\) 3.14  & +5.78 \(\pm\) 3.62 \\
\bottomrule
\end{tabular}
\caption{Attribute-related 2AFC gaps before and after balancing by OpenCLIP visual distance. Values are mean \(\pm\) population standard deviation across subjects, in percentage points. Positive values indicate higher 2AFC for the non-Attribute family.}
\label{tab:distance_balanced_family_gaps}
\end{table*}

\end{document}